\documentclass[sigconf]{acmart} 
\usepackage{graphicx}
\usepackage{amsmath}
\usepackage{subcaption}
\usepackage{epstopdf}
\usepackage{amsthm}
\usepackage{booktabs}
\usepackage{hyperref}
\usepackage{algorithm}
\usepackage{algorithmic}
\usepackage{multirow}

\usepackage{romannum}
 
\setcopyright{none}
 
\acmDOI{00.000/000_0}
 
\acmISBN{000-0000-00-000/00/00}
 
\acmConference[0000]{000}{0000}{Beijing, China} 
\acmYear{0000}
\copyrightyear{0000}
\acmPrice{00.00}

\begin{document}
	\title{A New Index for Clustering Evaluation Based on Density Estimation}
	
	\author{Gangli Liu}
	\affiliation{%
		\institution{Tsinghua University}
	}
	\email{gl-liu13@mails.tsinghua.edu.cn}

	\begin{abstract}
A new index for internal evaluation of clustering is introduced. The index is defined as a mixture of two sub-indices. The first sub-index $ I_a $ is called the Ambiguous Index; the second sub-index $ I_s $ is called the Similarity Index. Calculation of the two sub-indices is based on density estimation to each cluster of a partition of the data. An experiment is conducted to test the performance of the new index, and compared with six other internal clustering evaluation indices -- Calinski-Harabasz index, Silhouette coefficient, Davies-Bouldin index, CDbw,  DBCV, and VIASCKDE, on a set of 145 datasets. The result shows the new index significantly improves other internal clustering evaluation indices.

\end{abstract}
 
	\maketitle
 
\section{Introduction}

Cluster analysis or clustering is the task of grouping a set of objects in such a way that objects in the same group (called a cluster) are more similar (in some sense) to each other than to those in other groups (clusters). Due to its intensive applications in data analysis, people have developed a wide variety of  clustering algorithms. Such as K-Means, Gaussian mixtures, Spectral clustering \cite{ng2001spectral}, Hierarchical clustering \cite{johnson1967hierarchical}, DBSCAN \cite{schubert2017dbscan}, BIRCH \cite{zhang1996birch} etc.  With so many choices of clustering algorithms, it is vital to have good tools to recognize it when a clustering algorithm has done a good job.

Evaluation (or ``validation") of clustering results is as difficult as the clustering itself \cite{pfitzner2009characterization}. 
Popular approaches involve ``internal" evaluation and  ``external" evaluation.  In internal evaluation, a clustering result is evaluated based on the data that was clustered itself. Popular internal evaluation indices are Davies-Bouldin index \cite{ petrovic2006comparison}, Silhouette coefficient \cite{aranganayagi2007clustering}, Dunn index \cite{bezdek1995cluster}, and Calinski-Harabasz index  \cite{maulik2002performance} etc.   In external evaluation, the clustering result is compared to an existing ``ground truth" classification, such as the Rand index \cite{yeung2001details}.  However, ``knowledge of the ground truth classes is almost never available in practice'' \cite{scikit-learn}. In this article, we introduce a new index that belongs to the ``internal evaluation'' category of clustering evaluation, it is based on density estimation.

\section{RELATED WORK} \label{Sec_related}

Calinski-Harabasz index \cite{maulik2002performance}, Silhouette coefficient \cite{aranganayagi2007clustering}, and Davies-Bouldin index \cite{ petrovic2006comparison} are three of the most popular techniques for internal clustering evaluation.

\subsection{Calinski-Harabasz index (CH)}

For a set of data $  E $  of size  $ n_E $ which has been clustered into $ k $  clusters, the Calinski-Harabasz score $ s $ is defined as the ratio of the between-clusters dispersion mean and the within-cluster dispersion: \cite{scikit-learn}

\begin{equation*}
s = \frac{\mathrm{tr}(B_k)}{\mathrm{tr}(W_k)} \times \frac{n_E - k}{k - 1}
\label{eq:CH_index}
\end{equation*}

where $ {tr}(B_k) $  is trace of the between group dispersion matrix and $ {tr}(W_k) $ is the trace of the within-cluster dispersion matrix defined by:

\begin{equation*}
W_k = \sum_{q=1}^k \sum_{x \in C_q} (x - c_q) (x - c_q)^T
\end{equation*}

\begin{equation*}
B_k = \sum_{q=1}^k n_q (c_q - c_E) (c_q - c_E)^T
\end{equation*}

with  $ C_q $ the set of points in cluster $ q $, $ c_q $ the center of cluster $ q $,  $ c_E $ the center of $ E $ , and  $ n_q $ the number of points in cluster $ q $.

A higher Calinski-Harabasz score relates to a model with better defined clusters (Table \ref{tab:higher}).

\subsection{Silhouette coefficient (SC)}

The Silhouette coefficient for a single sample is given as:
\begin{equation*}
s = \frac{b - a}{max(a, b)}
\label{eq:S_index}
\end{equation*}
where $ a $ is the mean distance between a sample and all other points in the same class. $ b $ is the mean distance between a sample and all other points in the next nearest cluster. The Silhouette coefficient for a set of samples is given as the mean of Silhouette coefficient for each sample. The range of Silhouette coefficient is  $ [-1, 1]$. A higher Silhouette coefficient score relates to a model with better defined clusters.

\subsection{Davies-Bouldin index (DB)}

The Davies-Bouldin index can be calculated by the following formula: \footnote{ \url{https://en.wikipedia.org/wiki/Cluster_analysis}}

\begin{equation*}
DB={\frac {1}{n}}\sum _{i=1}^{n}\max _{j\neq i}\left({\frac {\sigma _{i}+\sigma _{j}}{d(c_{i},c_{j})}}\right) 
\label{eq:DB_index}
\end{equation*}
where n is the number of clusters,  $ c_{i} $ is the centroid of cluster $ i $, $ \sigma _{i} $ is the average distance of all elements in cluster $ i $ to centroid $ c_{i} $, and $ d(c_{i},c_{j}) $ is the distance between centroids $ c_{i} $ and $ c_{j} $. The clustering algorithm that produces a collection of clusters with the smallest Davies-Bouldin index is considered the best algorithm based on this criterion.

\begin{table*}
	\small
    \centering
    	\scalebox{0.90}{ 
\begin{tabular}{@{}ccccccc@{}}
	\toprule
	CH     & SC     & DB      & CDbw   & DBCV   & VIASCKDE & The new index \\ \midrule
	higher & higher & smaller & higher & higher & higher   & smaller       \\ \bottomrule
\end{tabular}

		}
\caption{Higher is better or smaller is better?}

\label{tab:higher}
\end{table*}

\subsection{Other internal evaluation indices}
Besides the three popular internal clustering evaluation indices, numerous criteria have been proposed in the literature \cite{vendramin2010relative,chou2004new,vzalik2011validity,baya2013many,moulavi2014density}. Such as DBCV \cite{moulavi2014density}, Xie-Beni (XB) \cite{xie1991validity}, CDbw \cite{halkidi2008density}, S\_Dbw \cite{halkidi2001clustering}, and RMSSTD \cite{halkidi2001clustering2001}. Besides, new cluster validity indices keep emerging, such as the CVNN \cite{liu2013understanding}, CVDD \cite{hu2019internal}, DSI \cite{guan2020internal}, SCV \cite{xu2020efficient}, AWCD \cite{li2020new} and VIASCKDE \cite{csenol2022viasckde}.

It is impractical to compare with all the indices listed above. Therefore, we select six of above internal clustering evaluation indices -- Calinski-Harabasz index, Silhouette coefficient, Davies-Bouldin index, CDbw,  DBCV, and VIASCKDE, to compare with the new index (see Section \ref{sec:experi}). Among them, Calinski-Harabasz index, Silhouette coefficient,  and Davies-Bouldin index are well known popular internal indices. CDbw and DBCV are two popular density-based internal clustering evaluation indices. VIASCKDE is a newly published index.
 
\subsection{Kernel density estimation} 

In probability and statistics, density estimation is the construction of an estimate, based on observed data, of an unobservable underlying probability density function \cite{silverman2018density}. 

Kernel density estimation (KDE) is a non-parametric way to estimate the probability density function of a random variable. Kernel density estimation is a fundamental data smoothing problem where inferences about the population are made, based on a finite data sample. \footnote{ \url{https://en.wikipedia.org/wiki/Kernel_density_estimation}}

Let $ (x_1, x_2, ..., x_n) $ be independent and identically distributed samples drawn from some univariate distribution with an unknown density $ f $ at any given point $ x $. We are interested in estimating the shape of this function $ f $. Its kernel density estimator is:

\begin{equation*}
{\widehat {f}}_{h}(x)={\frac {1}{n}}\sum _{i=1}^{n}K_{h}(x-x_{i})={\frac {1}{nh}}\sum _{i=1}^{n}K{\Big (}{\frac {x-x_{i}}{h}}{\Big )}
\end{equation*}

where $ K $ is the kernel -- a non-negative function -- and $ h > 0 $ is a smoothing parameter called the bandwidth. A kernel with subscript $ h $ is called the scaled kernel and defined as $ K_{h}(x) = 1/h K(x/h) $.  A range of kernel functions are commonly used: uniform, triangular, biweight, triweight, Epanechnikov, normal, and others.


\section{Calculation of the new index}  \label{sec:cal_index}

\begin{definition}
	The new index
	
	Suppose dataset  $\Omega$  of size $n_\Omega$ has been clustered into $ K $  clusters, the new index $ I $ is defined as a mixture of two sub-indices:
	\begin{equation}
	I = \delta I_a + (1 -  \delta)I_s
	\label{equ:index_I}
	\end{equation}
\end{definition}
where $ \delta $ is a hyper-parameter in range $ [0,1] $ which can be optimized by a training dataset.
The first sub-index $ I_a $ is called the Ambiguous Index; the second sub-index $ I_s $ is called the Similarity Index.  The range of  the new index  $ I$ is $ [0,1] $, and it is ``the smaller, the better''.

To calculate the two sub-indices, we first fit a density estimator $ d_q $ for each cluster $ C_q $.

\subsection{ The Ambiguous Index} 
To calculate the Ambiguous Index $ I_a $, we first determine the Territory of Cluster $ q $. 

\begin{definition}
	Territory of Cluster $ q $
	
	The Territory of Cluster $ q $, noted $ T_{q} $, is calculated with following equations:
 
	\begin{equation}
	G_q =\{ log(d_q (x)) \mid  x \in  C_q \}
	\label{equ:G_q1}
	\end{equation}
 
	\begin{equation}
	T_{q} = [min( G_q )  - \alpha_1 * \Delta_G, \quad max( G_q )  + \alpha_2 *  \Delta_G]
	\label{equ:T_q}
	\end{equation}
\end{definition}
where $ d_q (x) $ is the likelihood of $ x $ estimated by density estimator $ d_q $. $ log(d_q (x)) $ is its log-likelihood. $ \Delta_G  $ is Standard Deviation (SD) of $ G_q $. $\alpha_1 $ and $ \alpha_2 $ are two constant hyper-parameters. If $  \Delta_G  $ equals zero, then replace $ \alpha_1 * \Delta_G  $ and $ \alpha_2 * \Delta_G  $ with two constants $ \beta_1 $ and $ \beta_2 $. The hyper-parameters can be optimized by a training dataset.

$ T_{q} $ is a range for log-likelihood. To determine if a data point $ y $ is in the Territory of Cluster $ q $, we calculate the  log-likelihood of $ y $ with the density estimator $ d_q $. If $ log(d_q (y)) $ is in the range of  $ T_{q} $, then $ y $ is in the Territory of Cluster $ q $.  

\begin{definition} \label{def:am_poi}
	Ambiguous Point	
	
	If a data point is in the territories of at least two clusters, it is defined as an Ambiguous Point. 
\end{definition}

\begin{figure}
	\centering
	\includegraphics[width=0.9\columnwidth]{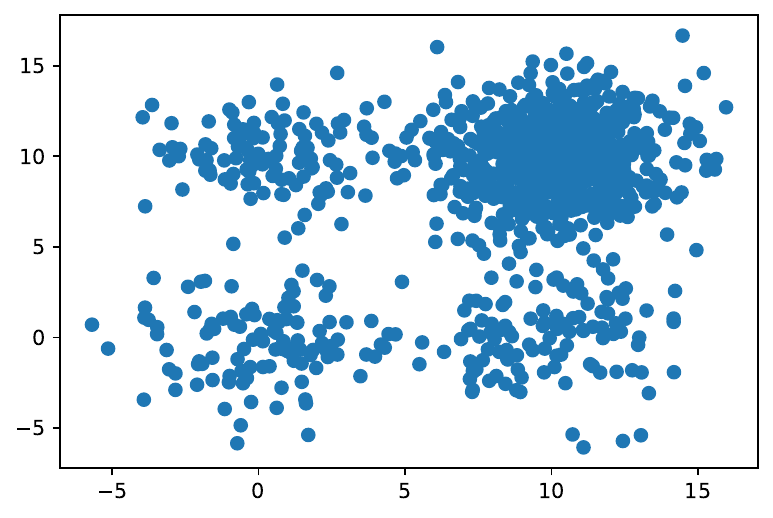}
	\caption{A dataset}
	\label{fig:data4}
\end{figure}

\begin{figure}
	\centering
	\includegraphics[width=0.9\columnwidth]{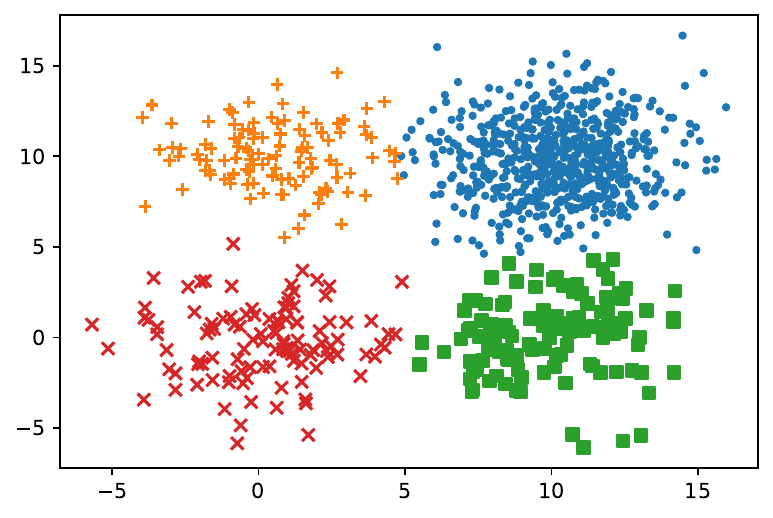}
	\caption{A partition of the dataset}
	\label{fig:data4_cluster}
\end{figure}

\begin{figure}
	\centering
	\includegraphics[width=0.9\columnwidth]{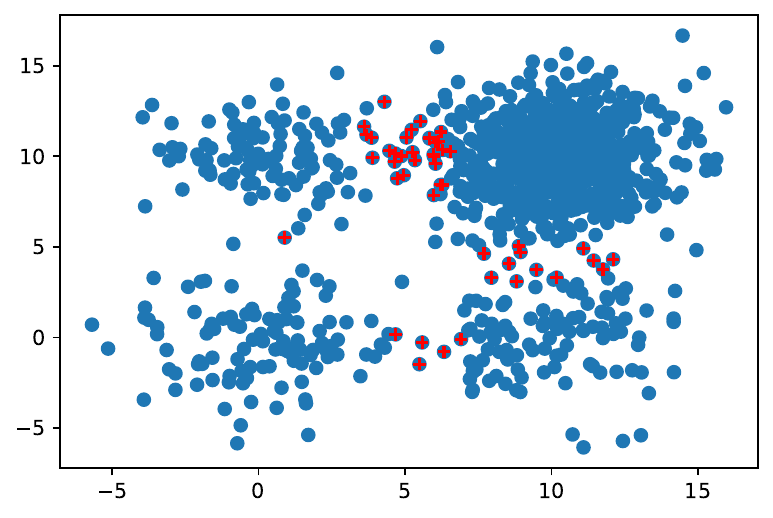}
	\caption{Ambiguous Points of the partition}
	\label{fig:data4_ambi}
\end{figure}

\begin{definition} 
	Ambiguous Index
	
	The Ambiguous Index is defined as the ratio of number of Ambiguous Points  in  $\Omega$ to total number of points in  $\Omega$.
	\begin{equation}
	I_{a} = \frac{a_\Omega}{n_\Omega}
	\end{equation}
\end{definition}
The range of  Ambiguous Index  $ I_{a} $ is $ [0,1] $.
Figure \ref{fig:data4}, \ref{fig:data4_cluster},  and	\ref{fig:data4_ambi} illustrate an example of calculating the Ambiguous Index. The dataset is from a previous paper \cite{handl2004multiobjective}. A clustering algorithm partitions the dataset into four clusters. The points tagged with a red cross in Figure \ref{fig:data4_ambi} are Ambiguous Points, according to Definition  \ref{def:am_poi}.

\begin{table*}
	\begin{center}
		\scalebox{0.8}{
	
 \begin{tabular}{@{}ll@{}}
 	\toprule
 	& URL                                                                             \\
 	\midrule
 	CDbw     & \url{https://pypi.org/project/cdbw/}                                                 \\
 	DBCV     & \url{https://github.com/scikit-learn-contrib/hdbscan/blob/master/hdbscan/validity.py} \\
 	VIASCKDE & \url{https://github.com/senolali/VIASCKDE}      \\ \bottomrule                                     
 \end{tabular}

	}		
	
		\caption{URL of implementations of three density-based indices}
		\label{tab:three_den}
	\end{center}
\end{table*}

\subsection{ The Similarity Index} 
The Similarity Index is calculated with the following equations:
\begin{definition} 
	Similarity Index
	
	\begin{equation}
	L_q =\{ d_q (x) \mid  x \in  C_q \}
	\label{equ:L_q}
	\end{equation}
	
	\begin{equation}
	S_q =  \frac {1}{max(L_q)} \sum _{y\in  L_q } \; y  
	\label{equ:S_q}
	\end{equation}

	\begin{equation}
	S_\Omega = \sum_{q = 1 } ^{K }\; S_q
	\end{equation}
	
	\begin{equation}
	I_{s} = 1 - \frac{S_\Omega }{n_\Omega}
	\end{equation}
	where $ d_q (x) $ is the likelihood of data point $ x $ evaluated by density estimator $ d_q $, $ L_q $  is the set of all points' likelihood in cluster $ C_q $. $ max(L_q) $ is the maximum value in $ L_q $. $ S_q $ is sum of likelihood normalized by the maximum likelihood in cluster $ C_q $.  $ S_\Omega $ is sum of all normalized likelihood in data $ \Omega $. $ n_\Omega $ is the number of points in data  $\Omega $. $ \frac{S_\Omega }{n_\Omega} $ is the average normalized likelihood in  $ \Omega $. Note the difference between likelihood and log-likelihood. Here we use likelihood.
\end{definition}
The range of  Similarity Index  $ I_s$ is also $ [0,1] $.

\begin{figure}[!h]
	\centering
	\includegraphics[width=0.7\columnwidth]{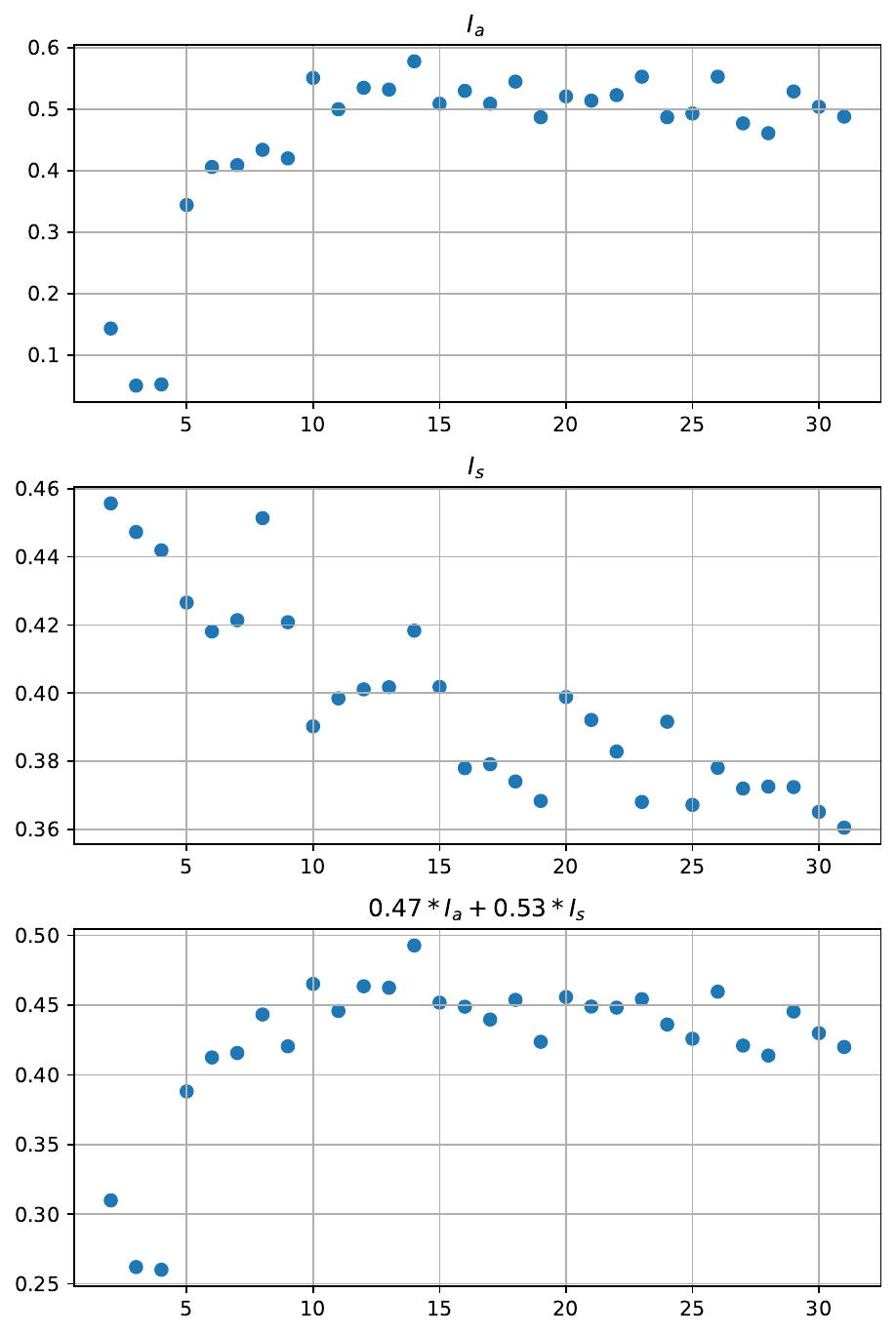}
	\caption{Mixture of the two sub-indices works}
	\label{fig:4_clu_3_index}
\end{figure}
 
\section{Experiment}  \label{sec:experi}
We conducted an experiment to test the performance of the new index. 

\subsection{A set of datasets}  \label{sec:data_145}
In the experiment, we collected a set of 145 datasets from two benchmarks of evaluating clustering (see Table \ref{tab:two_bench}) and the scikit-learn project \cite{scikit-learn}. All the datasets are  two-dimensional or three-dimensional, so that it is convenient to validate the outcomes by eyeballing it.

\begin{table}
\begin{center}
\scalebox{0.8}{

\begin{tabular}{ll}
\toprule
{} 	& URL    \\
\midrule
Benchmark1 & \url{https://github.com/deric/clustering-benchmark}        \\
Benchmark2 & \url{https://github.com/gagolews/clustering\_benchmarks\_v1}  \\
\bottomrule
\end{tabular} 

}
\caption{Two benchmarks for evaluating clustering}
\label{tab:two_bench}
\end{center}
\end{table}

\subsection{Settings of the experiment} 
For each dataset, seven clustering algorithms are used to partition the dataset with parameter K (number of clusters) ranging from 2 to 30. The seven clustering algorithms are: Agglomerative Clustering with ward, complete,  average, and single linkage (each linkage is counted as a different clustering algorithm), Spectral Clustering, K-means Clustering, and Gaussian Mixture Clustering. 

If two clustering algorithms' partitioning of a dataset are the same, only one partitioning is kept. In addition, a reference partition (it is supposed to be the right partition) is provided for each dataset (see the first row of Figure \ref{fig:res_145_4}).

Therefore, if there are no duplicate partitions and the reference partition is not in the seven clustering algorithms' outcomes, then there are 204 (29 * 7  + 1 = 204) candidate partitions of a dataset to be evaluated by an index.

We used about 1/4 of the datasets (that is 36 of the 145 datasets)  to select  optimized values for the hyper-parameters in Equation \ref{equ:index_I} and \ref{equ:T_q}. For density estimation of a cluster, we used the Kernel Density Estimation (KDE) library functions provided by the scikit-learn project. With the ``bandwidth'' hyper-parameter selected by Cross Validation and Grid Search, and the ``kernel'' fixed to be \emph{``gaussian''}.

Then, we used the new index to evaluate the 204 (may be less due to duplicates) candidate partitions of each dataset, selecting the top 5 partitions ranked by the index. Then compared the results with the results evaluated by the six indices mentioned in Section \ref{Sec_related}. Scores of Calinski-Harabasz index, Silhouette coefficient, and Davies-Bouldin index are calculated with library functions provided by the scikit-learn project. Implementations of CDbw,  DBCV, and VIASCKDE see Table \ref{tab:three_den}.

\subsection{Results of the experiment} 
\begin{figure*}
	\centering
	\includegraphics[width=1.5\columnwidth]{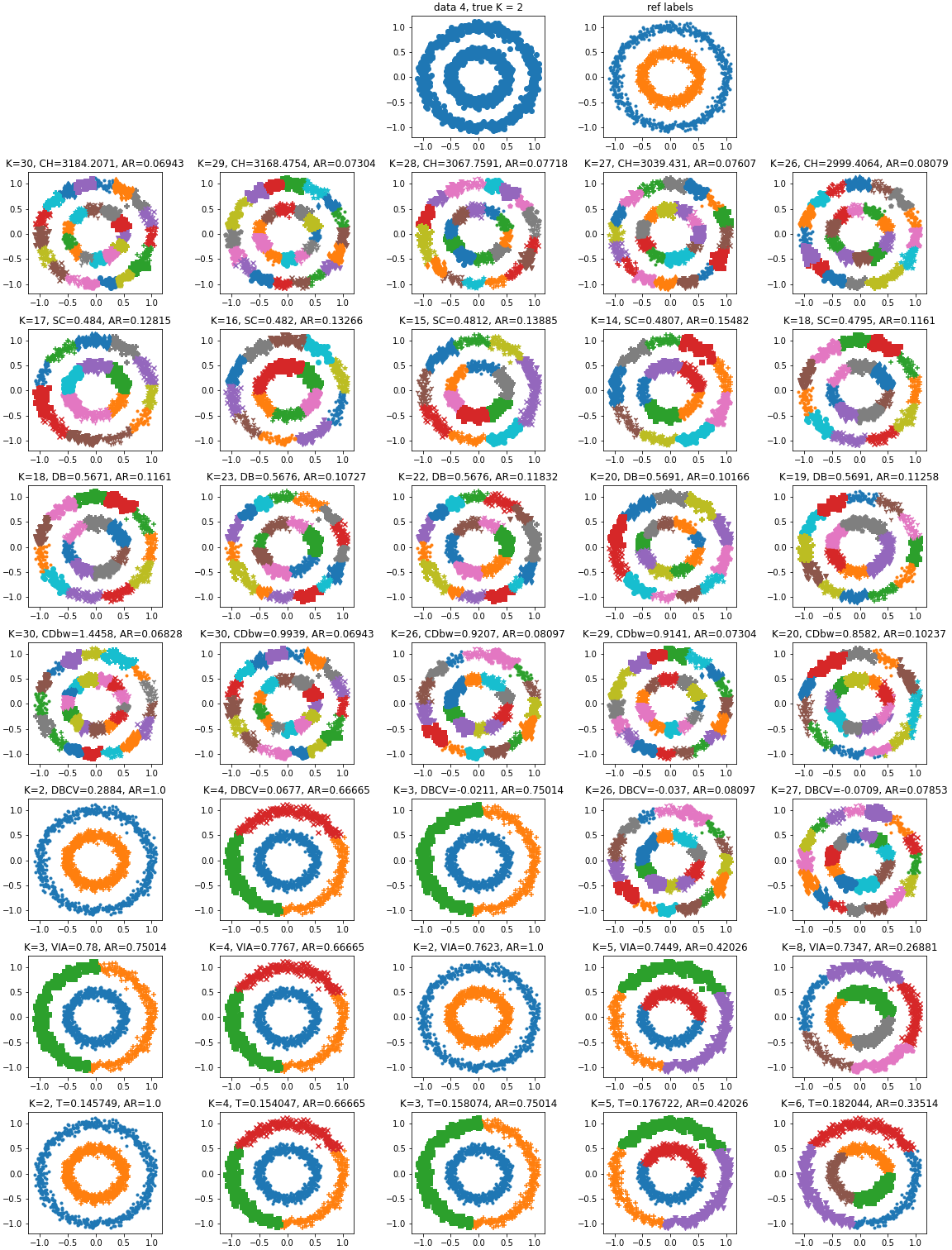}
	\caption{Result of one dataset}
	\label{fig:res_145_4}
\end{figure*}

Figure \ref{fig:res_145_4} is result for one dataset.  The first row of the figure plots the dataset, tagged with dataset ID and the true number of clusters K, and the reference partition of the dataset. The following seven rows plot the top 5 partitions evaluated by each index (ranked from left to right). Each tagged with number of clusters K and scoring of the corresponding index, and the adjusted Rand index (AR) of the partition with the reference partition.  The adjusted Rand index is used to evaluate a partition's agreement with the reference partition of the dataset. ``CH'' stands for Calinski-Harabasz index, ``SC'' stands for Silhouette coefficient, ``DB'' stands for Davies-Bouldin index, and ``T'' is the new index. ``VIASCKDE'' is simplified to ``VIA''.

Note the Davies-Bouldin index and the new index are ``the smaller, the better''; while others are ``the larger, the better''. It can be seen from Figure \ref{fig:res_145_4} that some of the indices has recognized the best partition of the data, while others failed.

\begin{table*}
	\small
    \centering
    	\scalebox{0.80}{
           \begin{tabular}{c|c|c|c|c|c|c|c|c|c|c|c|c|c|c|c|c|c|c|c|c|c|c|c|c|c}
    		\hline
    	~ & 0   & 1   & 2   & 3   & 4   & 5   & 6   & 7   & 8   & 9   & 10  & 11  & 12  & 13  & 14  & 15  & 16  & 17  & 18  & 19  & 20  & 21  & 22  & 23  & 24  \\ \hline 
    	CH & S  & S  & F  & F  & F  & F  & F  & F  & F  & F  & F  & S  & S  & S  & F  & F  & F  & S  & S  & F  & F  & F  & F  & F  & F   \\ \hline 
    	SC & S  & S  & F  & F  & F  & F  & F  & F  & S  & F  & F  & S  & S  & S  & S  & F  & F  & S  & F  & F  & F  & F  & F  & F  & F  \\ \hline 
    	DB & S  & S  & F  & F  & F  & F  & F  & F  & S  & F  & F  & S  & S  & S  & S  & F  & F  & S  & F  & S  & F  & F  & F  & F  & F \\ \hline
    	CDbw & S  & F  & F  & F  & F  & F  & F  & F  & F  & F  & F  & F  & F  & S  & F  & F  & F  & F  & F  & F  & F  & F  & F  & F  & F    \\ \hline 
    	DBCV & F  & F  & F  & S  & S  & F  & F  & S  & S  & F  & F  & S  & S  & S  & S  & F  & S  & S  & F  & S  & F  & F  & S  & F  & S   \\ \hline 
    	VIA & F  & F  & F  & F  & F  & F  & F  & F  & F  & F  & S  & F  & F  & F  & F  & F  & F  & F  & F  & F  & F  & F  & F  & F  & F   \\ \hline 
    	New & S  & S  & S  & S  & S  & S  & S  & F  & S  & S  & S  & S  & S  & S  & S  & S  & S  & S  & F  & F  & F  & F  & F  & S  & F   \\ \hline

    	& 25  & 26  & 27  & 28  & 29  & 30  & 31  & 32  & 33  & 34  & 35  & 36  & 37  & 38  & 39  & 40  & 41  & 42  & 43  & 44  & 45  & 46  & 47  & 48  & 49  \\ \hline 
    	CH & F  & F  & F  & S  & F  & S  & F  & F  & F  & F  & F  & F  & F  & F  & F  & F  & F  & F  & F  & F  & F  & F  & F  & F  & F   \\ \hline 
    	SC & F  & F  & F  & S  & F  & S  & F  & F  & F  & F  & F  & F  & F  & F  & S  & F  & S  & F  & F  & F  & F  & F  & F  & F  & F  \\ \hline 
    	DB & S  & F  & F  & S  & F  & S  & F  & F  & S  & F  & F  & F  & F  & F  & S  & F  & S  & S  & F  & F  & F  & F  & F  & F  & F \\ \hline
    	CDbw & F  & F  & F  & F  & S  & F  & F  & F  & S  & F  & F  & F  & F  & F  & F  & F  & F  & F  & S  & S  & F  & F  & F  & F  & F    \\ \hline 
    	DBCV & F  & F  & S  & F  & F  & S  & F  & S  & F  & F  & F  & F  & F  & F  & S  & F  & S  & S  & S  & S  & F  & F  & S  & F  & F   \\ \hline 
    	VIA & F  & F  & S  & F  & F  & F  & F  & S  & S  & F  & F  & F  & F  & F  & F  & F  & F  & F  & F  & F  & F  & F  & F  & F  & F   \\ \hline 
    	New & S  & S  & S  & F  & S  & S  & S  & S  & S  & S  & S  & F  & F  & F  & S  & F  & F  & S  & F  & S  & S  & F  & S  & F  & S   \\ \hline

    	& 50  & 51  & 52  & 53  & 54  & 55  & 56  & 57  & 58  & 59  & 60  & 61  & 62  & 63  & 64  & 65  & 66  & 67  & 68  & 69  & 70  & 71  & 72  & 73  & 74  \\ \hline 
    	CH & S  & F  & F  & F  & F  & S  & F  & F  & F  & F  & F  & F  & F  & F  & F  & F  & F  & S  & F  & S  & F  & F  & F  & F  & F   \\ \hline 
    	SC & S  & F  & S  & F  & F  & S  & F  & F  & F  & F  & F  & F  & F  & F  & F  & F  & F  & S  & F  & S  & S  & F  & F  & F  & S  \\ \hline 
    	DB & S  & F  & S  & F  & F  & S  & F  & F  & F  & F  & F  & F  & F  & F  & F  & F  & F  & S  & F  & S  & S  & F  & F  & F  & S \\ \hline
    	CDbw & F  & F  & F  & F  & F  & F  & F  & F  & F  & F  & F  & F  & F  & F  & F  & F  & F  & F  & F  & F  & F  & F  & F  & F  & F    \\ \hline 
    	DBCV & S  & F  & F  & F  & F  & S  & F  & F  & F  & F  & S  & F  & F  & F  & F  & S  & F  & F  & S  & S  & S  & F  & F  & F  & S   \\ \hline 
    	VIA & F  & F  & F  & F  & F  & S  & F  & F  & F  & F  & F  & F  & F  & F  & F  & S  & F  & F  & F  & F  & F  & S  & F  & F  & F   \\ \hline 
    	New & F  & F  & S  & F  & F  & S  & F  & S  & F  & S  & S  & S  & F  & F  & S  & F  & S  & F  & S  & S  & S  & F  & F  & F  & S   \\ \hline

    	& 75  & 76  & 77  & 78  & 79  & 80  & 81  & 82  & 83  & 84  & 85  & 86  & 87  & 88  & 89  & 90  & 91  & 92  & 93  & 94  & 95  & 96  & 97  & 98  & 99  \\ \hline 
    	CH & F  & F  & F  & S  & F  & F  & F  & F  & S  & F  & F  & S  & F  & F  & F  & F  & F  & F  & F  & F  & F  & F  & S  & F  & S   \\ \hline 
    	SC & S  & F  & F  & S  & F  & S  & S  & F  & S  & F  & S  & S  & F  & S  & F  & F  & S  & S  & F  & F  & F  & F  & S  & F  & S  \\ \hline 
    	DB & S  & F  & F  & S  & F  & S  & S  & F  & S  & F  & S  & S  & F  & F  & F  & F  & S  & S  & F  & F  & F  & F  & S  & F  & S \\ \hline
    	CDbw & F  & F  & F  & F  & F  & F  & F  & F  & F  & F  & F  & F  & F  & F  & F  & F  & F  & F  & F  & F  & S  & F  & F  & F  & F    \\ \hline 
    	DBCV & S  & F  & F  & S  & F  & S  & S  & F  & F  & F  & S  & S  & S  & F  & F  & F  & S  & S  & F  & F  & S  & F  & S  & F  & S   \\ \hline 
    	VIA & F  & F  & F  & F  & F  & F  & F  & F  & S  & F  & F  & F  & F  & F  & F  & F  & F  & F  & F  & F  & F  & F  & F  & F  & F   \\ \hline 
    	New & S  & F  & F  & F  & S  & S  & F  & F  & S  & F  & S  & F  & S  & F  & F  & F  & S  & S  & F  & S  & S  & F  & S  & S  & F   \\ \hline

    	& 100 & 101 & 102 & 103 & 104 & 105 & 106 & 107 & 108 & 109 & 110 & 111 & 112 & 113 & 114 & 115 & 116 & 117 & 118 & 119 & 120 & 121 & 122 & 123 & 124 \\ \hline 
    	CH & S  & F  & F  & S  & F  & F  & F  & F  & S  & S  & F  & F  & F  & F  & F  & F  & F  & F  & F  & F  & F  & F  & F  & S  & F   \\ \hline 
    	SC & S  & F  & F  & S  & F  & F  & F  & F  & S  & F  & F  & F  & F  & F  & F  & F  & F  & F  & F  & F  & F  & F  & F  & S  & F  \\ \hline 
    	DB & S  & F  & F  & S  & F  & F  & F  & F  & S  & S  & F  & F  & F  & F  & F  & F  & F  & F  & F  & F  & F  & F  & F  & F  & F \\ \hline
    	CDbw & F  & F  & F  & F  & F  & F  & F  & F  & F  & F  & F  & F  & F  & F  & S  & F  & F  & F  & F  & F  & F  & F  & F  & F  & F    \\ \hline 
    	DBCV & F  & F  & F  & S  & S  & F  & F  & F  & S  & F  & F  & S  & F  & S  & S  & F  & F  & F  & S  & F  & F  & S  & F  & F  & F   \\ \hline 
    	VIA & F  & F  & F  & F  & S  & F  & F  & S  & F  & F  & F  & F  & F  & F  & F  & F  & F  & F  & F  & F  & F  & F  & F  & F  & F   \\ \hline 
    	New & S  & F  & F  & S  & S  & F  & F  & F  & S  & S  & F  & S  & F  & F  & F  & F  & F  & S  & S  & S  & S  & S  & F  & F  & F   \\ \hline

    	& 125 & 126 & 127 & 128 & 129 & 130 & 131 & 132 & 133 & 134 & 135 & 136 & 137 & 138 & 139 & 140 & 141 & 142 & 143 & 144 &     &     &     &     &     \\ \hline 
    	CH & F  & S  & F  & F  & S  & F  & F  & F  & F  & F  & F  & F  & F  & F  & S  & F  & F  & F  & F  & S    &     &     &     &     &     \\ \hline 
    	SC & F  & F  & F  & F  & S  & F  & F  & F  & F  & F  & F  & F  & F  & F  & S  & F  & F  & F  & F  & S    &     &     &     &     &  \\ \hline 
    	DB & F  & F  & F  & F  & S  & F  & F  & F  & F  & S  & F  & F  & F  & F  & S  & F  & F  & F  & F  & S & & & & & \\ \hline
    	CDbw & F  & F  & F  & F  & F  & F  & F  & F  & F  & F  & F  & F  & F  & F  & F  & F  & F  & F  & F  & F & & & & &    \\ \hline 
    	DBCV & F  & F  & S  & S  & F  & F  & F  & F  & F  & S  & F  & S  & F  & F  & S  & F  & F  & F  & S  & F  & & & & &    \\ \hline 
    	VIA & F  & F  & F  & F  & F  & F  & F  & F  & F  & F  & F  & F  & F  & F  & F  & F  & F  & F  & F  & S   & & & & &  \\ \hline 
    	New & F  & F  & F  & F  & F  & F  & F  & F  & F  & F  & F  & F  & F  & S  & S  & F  & F  & S  & S  & S  & & & & &  \\ \hline

     \end{tabular}

		}
\caption{The seven indices' performance on the 145 datasets. `S': Succeeded, `F': Failed, `New': the new index}

\label{tab:big_table}
\end{table*}

People may have different opinion about what is the right partition to some datasets. Therefore, we present each index's scoring to different partitions to all the 145 datasets, so that people can make their own evaluation about the performance of the indices. The Appendix section lists 20 more samples of the results. A full list of all the 145 datasets are attached as auxiliary files of the paper.

\subsubsection{Counting accuracy of the indices} 
To count the accuracy of each index, we calculate the adjusted Rand index of the partition that is ranked first (the champion), with the reference partition. If it is greater than $0.95$, then an index is classified as ``Succeeded'' in recognizing the right partition. Otherwise, it is classified as ``Failed''. E.g., in the second row of Figure \ref{fig:res_145_4}, the adjusted Rand index of the champion with the reference is $0.06943$, which is less than $0.95$. Therefore, the CH index is classified as ``Failed'' for this dataset. While the sixth row, $AR = 1.0$, which is  greater than $0.95$, thus the DBCV index is classified as ``Succeeded'' in recognizing the right partition.

Table \ref{tab:big_table} shows the performance of the seven indices. Table \ref{tab:Accu} lists accuracy of the indices. According to the accuracy, we can conclude the new index has significantly outperformed other indices.
 
In the experiment, we select a best ``bandwidth" for each cluster $ C_q $. In a recent revision of the new index, we select a global best ``bandwidth" for the whole data  $ \Omega$, then use the  global best ``bandwidth" for all clusters. We also use a technique called ``sliding window" to select the global best ``bandwidth" from a large span. For details of  ``sliding window" see the Python code of Figure \ref{fig:sliding} in Appendix section. These two revisions improve the performance of the new index from $ 74/145 $ to $ 86/145 $.

In the implementation of VIASCKDE (Table \ref{tab:three_den}), the author used a fixed ``bandwidth'' of $ 0.05 $. If we select the ``bandwidth'' with Cross Validation and Grid Search, the performance of VIASCKDE can be improved from $ 11/145 $ to $ 21/145 $.

\begin{table*}
	\begin{center}
		\scalebox{0.8}{
	
	\begin{tabular}{llllllll}
		\toprule
		{}    &  CH   & SC    &  DB   &  CDbw   &   DBCV   &   VIASCKDE   & New  \\
		\midrule
		Accuracy & 27/145 & 38/145 & 42/145  & 8/145  & 56/145  & 11/145 & 74/145   \\
		\bottomrule
\end{tabular} 		}		
	
		\caption{Accuracy of the seven indices}
		\label{tab:Accu}
	\end{center}
\end{table*}

\section{Why it works?} 
\begin{figure*}
	\centering
	\includegraphics[width=1.5\columnwidth]{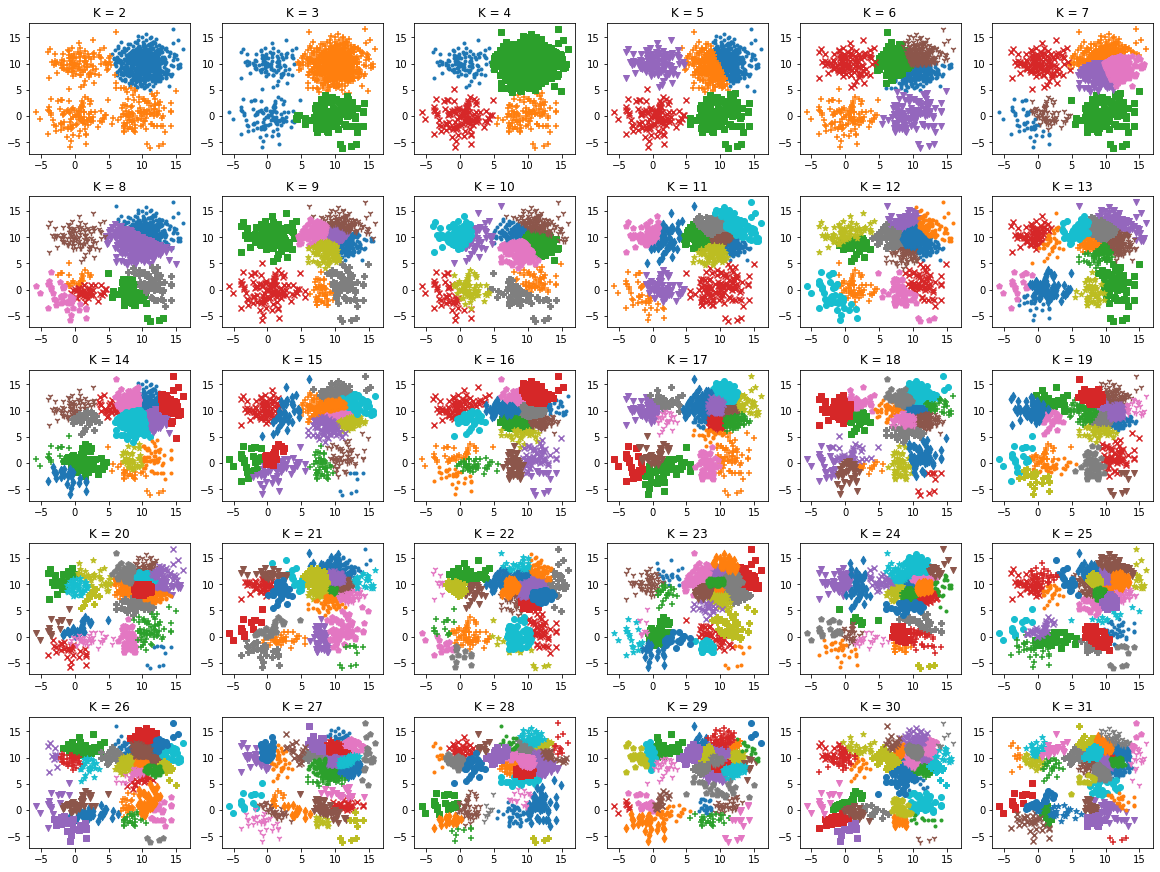}
	\caption{30 candidate partitions of a dataset}
	\label{fig:4_clu_30}
\end{figure*}

We use an example to intuitively show why the new index works. In the example, we use the Gaussian Mixture Model (GMM) to partition a dataset into, from $ K = 2 $ to $ K = 31 $ clusters, as illustrated in Figure \ref{fig:4_clu_30}.  

Figure \ref{fig:4_clu_3_index} shows how the two sub-indices and their sums varying along the number of clusters K. As mentioned in Section  \ref{sec:cal_index}, the range of  the new index  $ I$ is $ [0,1] $, and it is ``the smaller, the better''. To get a smaller index  $ I$,  the Ambiguous Index $ I_a $ tries to pull the number of clusters K to a smaller number; however, the Similarity Index $ I_s $ tries to pull the number of clusters K to a larger number. At some point, the two sub-indices get an equilibrium. The hyper-parameter in Equation \ref{equ:index_I} is to adjust the two sub-indices so that they are on comparable level.

\begin{figure*}
	\centering
	\includegraphics[width=1.5\columnwidth]{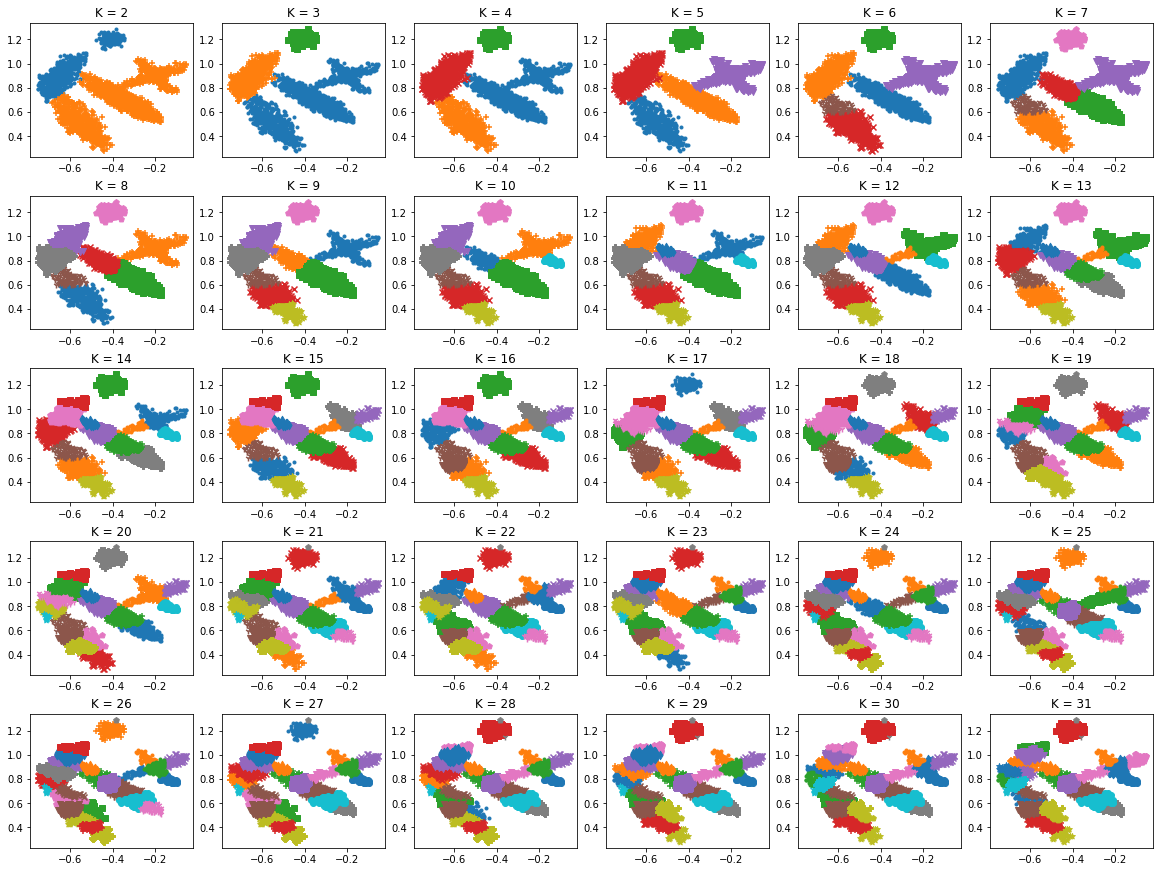}
	\caption{30 candidate partitions of another dataset}
	\label{fig:data_71_30_partitions}
\end{figure*}

\begin{figure*}
	\centering
	\includegraphics[width=1.8\columnwidth]{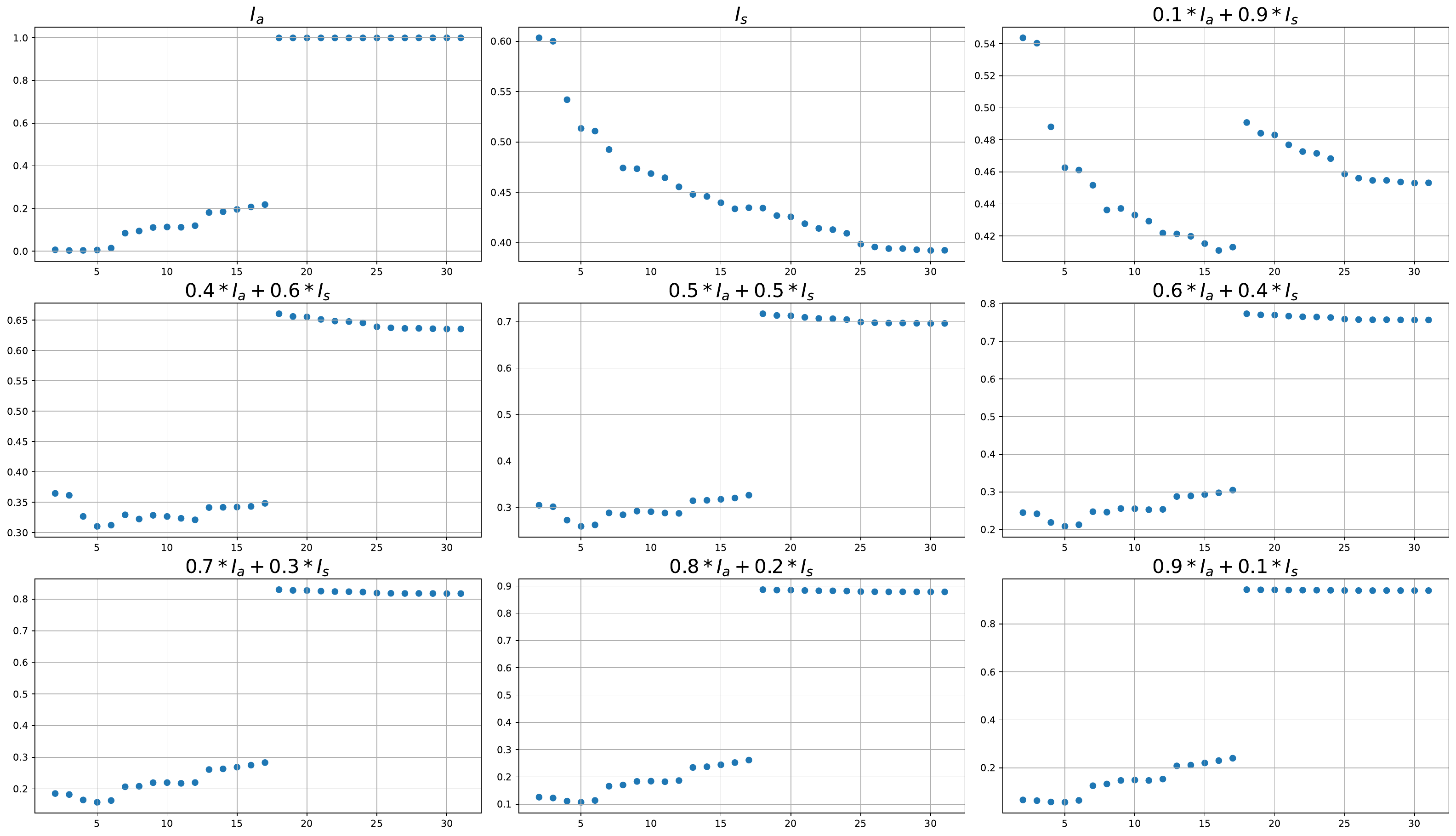}
	\caption{Effect of hyper-parameter $\delta$ in Equation \ref{equ:index_I}}
	\label{fig:data_71_30_parti_3_indices}
\end{figure*}

Another example is illustrated in Figure  \ref{fig:data_71_30_partitions} and \ref{fig:data_71_30_parti_3_indices}. Figure  \ref{fig:data_71_30_parti_3_indices} also shows how the result is affected by hyper-parameter $\delta$ in Equation \ref{equ:index_I}. It can be seen that the result is not very sensitive to hyper-parameter $\delta$. When $\delta$ locates in an appropriate range, the new index will succeed in recognizing the right partition of the data.

An interesting question is: what if we use only one sub-index? For some easy recognition tasks, one sub-index alone can accomplish the recognition of the right partition; for some harder tasks, we need the cooperation of the two sub-indices.

\section{Discussion} 

\subsection{Several variants of the two sub-indices} 
There are several variants of the two sub-indices, which get comparable performance scores to the main  sub-indices defined in Section \ref{sec:cal_index}, when tested on the datasets mentioned in Section \ref{sec:data_145}.

\subsubsection{Variant \Romannum{1} of the Ambiguous Index, $ I_{a\_v1} $} ~\\
To calculate $ I_{a\_v1} $, we first calculate the pairwise Ambiguous Index between each pair of clusters $ i $ and $ j $, noted $ A_{ij }$, then compute the proportion of $ A_{ij} $ that is positive.

\subsubsection{Variant \Romannum{2} of the Ambiguous Index, $ I_{a\_v2} $} ~\\
In $ I_{a\_v2} $, instead of computing the proportion of $ A_{ij} $ that is positive, we compute the mean of $ A_{ij} $ that is positive. If all $ A_{ij} $ are zero, we just set the value to zero.

\subsubsection{Variant \Romannum{3} of the Ambiguous Index, $ I_{a\_v3} $} ~\\
To calculate $ I_{a\_v3} $, we first estimate the area of the territory for each cluster by using Monte Carlo method, then calculate the proportion of the area that has disputes to the total area of all clusters.

\subsubsection{Variant \Romannum{1} of the Similarity Index, $ I_{s\_v1} $} ~\\
To calculate $ I_{s\_v1} $, we first process $ L_q $ in Equation 	\ref{equ:L_q}  with a min-max normalization, then calculate the $ S_q $ as the sum of the min-max normalized  $ L_q $.

\subsubsection{Variant \Romannum{2} of the Similarity Index, $ I_{s\_v2} $} ~\\
To calculate $ I_{s\_v2} $, we first calculate the center of $ G_q$ in Equation \ref{equ:G_q1}. It can be the mean or median of $ G_q$. Then calculate the distance (using absolute value) or squared-distance from each member of $ G_q$ to the center. Then calculate mean of the distances. The log-likelihood can be replaced with likelihood. The distances can be min-max normalized.

\subsection{The third sub-index, Boundary Index, $ I_b $}  
\begin{figure*} 
	\begin{subfigure}{0.4\textwidth}
	\includegraphics[width=\linewidth]{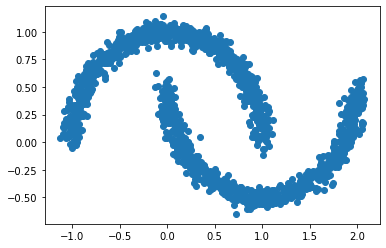}
\end{subfigure}    
\begin{subfigure}{0.4\textwidth}
	\includegraphics[width=\linewidth]{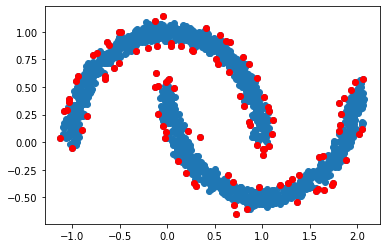}
\end{subfigure}    
\caption{Boundary Points} \label{fig:bounp}
\end{figure*}

To calculate Boundary Index $ I_b $, we first determine the Boundary Points of cluster $ C_q $.
\begin{definition} \label{def:boun_poi}
	Boundary Point	
	
	\begin{equation}
	B_{q} = [min( G_q ), \quad min( G_q )  + \rho *  \Delta_G]
	\label{equ:B_q}
	\end{equation}
	where $ G_q $ is defined in Equation \ref{equ:G_q1}, $ \Delta_G  $ is Standard Deviation (SD) of $ G_q $, $\rho$ is a constant hyper-parameter.
	If a data point $ p $'s log-likelihood evaluated by density estimator $ d_q $, is in the range of $ B_{q} $, then $ p $ is defined as a Boundary Point of cluster $ C_q $.  
\end{definition}
Figure \ref{fig:bounp} shows two clusters of data points with Boundary Points colored red.

\begin{definition} 
	Boundary Index
	\begin{equation}
	I_{b} = \frac{\sum _{i=1}^{K}N_{B_{i}}}{KN_\Omega}
	\end{equation}
	where $ N_{B_{i}}  $ is the number of Boundary Points in cluster $ i $, $ N_{\Omega} $ is the total number of points in data $ \Omega $, $ K $ is the number of clusters.
\end{definition}
Preliminary tests show a mixture of $ I_a $ and $ I_b $ also get a comparable performance score to the mixture of $ I_a $ and $ I_s $, when tested on the datasets mentioned in Section \ref{sec:data_145}.

\subsection{Other density estimation} 
It seems a good density estimator for each cluster $ C_q $ is critical to the success of the new index. In the experiment, we used kernel density estimation implemented by the scikit-learn project to obtain a density estimator for each cluster. It is possible to improve performance by using other density estimation models or implementations, such as Bayesian Gaussian Mixture.
 
\subsection{Mixture of more ingredients} 
It is possible to use a mixture of more ingredient indices to get a better performance. E.g., some preliminary tests show a mixture of $ I_a $, $ I_s $, and $ I_b$ can get a slightly better performance. However, the improvement is not significant.  Besides, it increases model complexity.

\subsection{Estimating the $\delta$ hyper-parameter} 
Maybe we should not fix the $\delta$ hyper-parameter in Equation \ref{equ:index_I} for all datasets, but to estimate it with characteristics of data $\Omega$. Such as number of points in $\Omega$, dimension of $\Omega$, global best bandwidth of $\Omega$ etc. However, how to calculate the $\delta$ need to be figured out.

\subsection{Complexity} 
The complexity of the new index depends on what density estimation techniques are employed,  to get a density estimator for each cluster $ C_q $. It is generally slower than computation of 
non-density-based indices. Because we need to fit a density estimator for each cluster firstly. The computation can be accelerated by employing parallel computing.

\subsection{Fixing random seeds} 
If randomness is involved in the process of fitting density estimators for each cluster,  or selecting the best bandwidth for dataset $\Omega$, the value of the new index can be fixed by fixing the random seeds.

\section{Conclusion and Future Works} 
We introduced a new index for internal evaluation of clustering, then tested and compared it with six popular or new internal clustering evaluation indices -- Calinski-Harabasz index, Silhouette coefficient, Davies-Bouldin index, CDbw,  DBCV, and VIASCKDE, on a set of 145 datasets collected from two benchmarks of clustering and the scikit-learn project. The result shows the new index significantly improves the six internal clustering evaluation indices. Further research will check whether the new index could work in higher dimension, and whether it could deal with noise or not.
 
\bibliographystyle{ACM-Reference-Format}
\bibliography{new_index}


\section{\\Appendix}

\appendix
\begin{figure*}
	\centering
	\includegraphics[width=1.5\columnwidth]{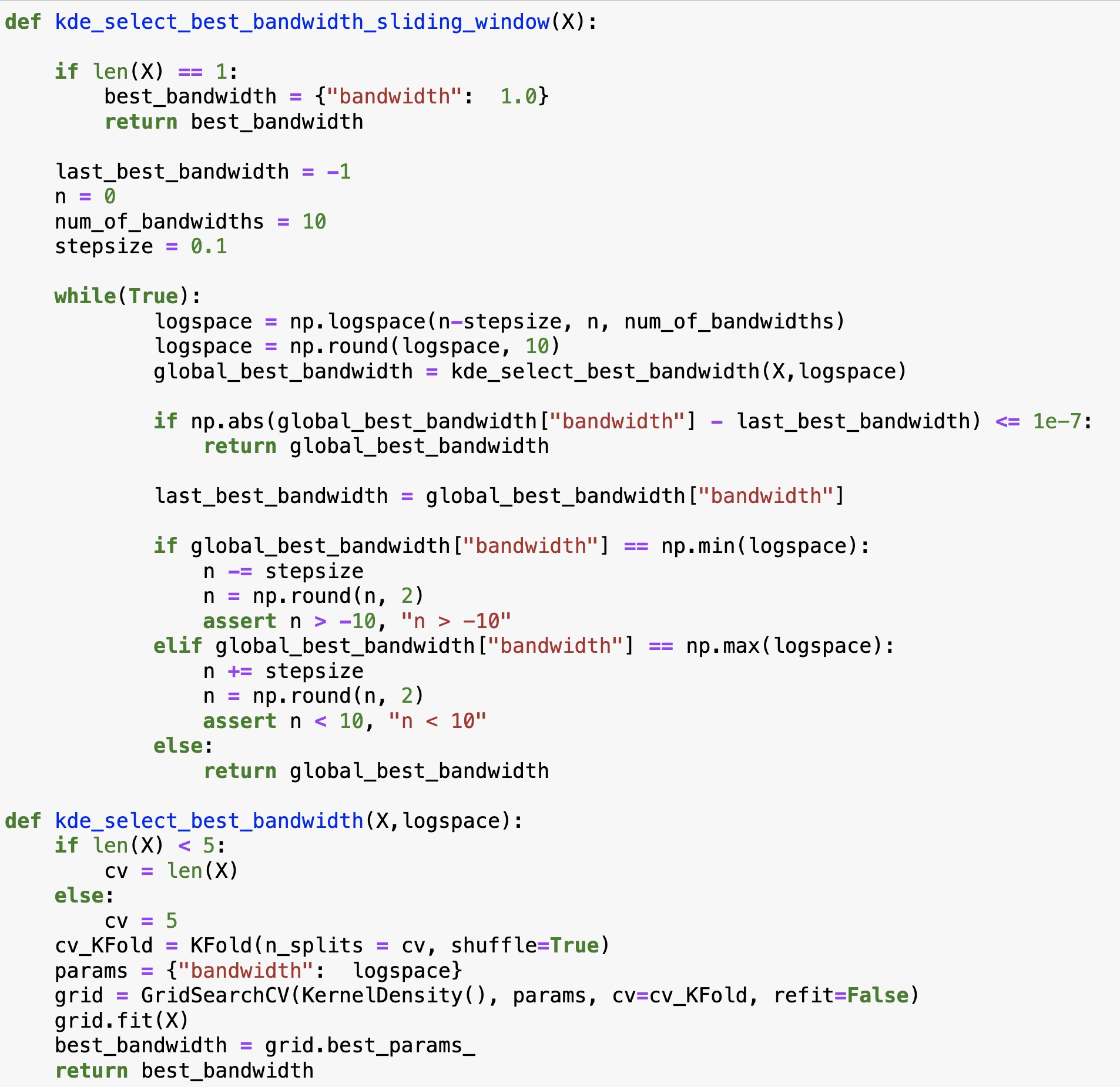}
	\caption{Python code of sliding window technique} \label{fig:sliding}
\end{figure*}

\begin{figure*}
	\centering
	\includegraphics[width=1.5\columnwidth]{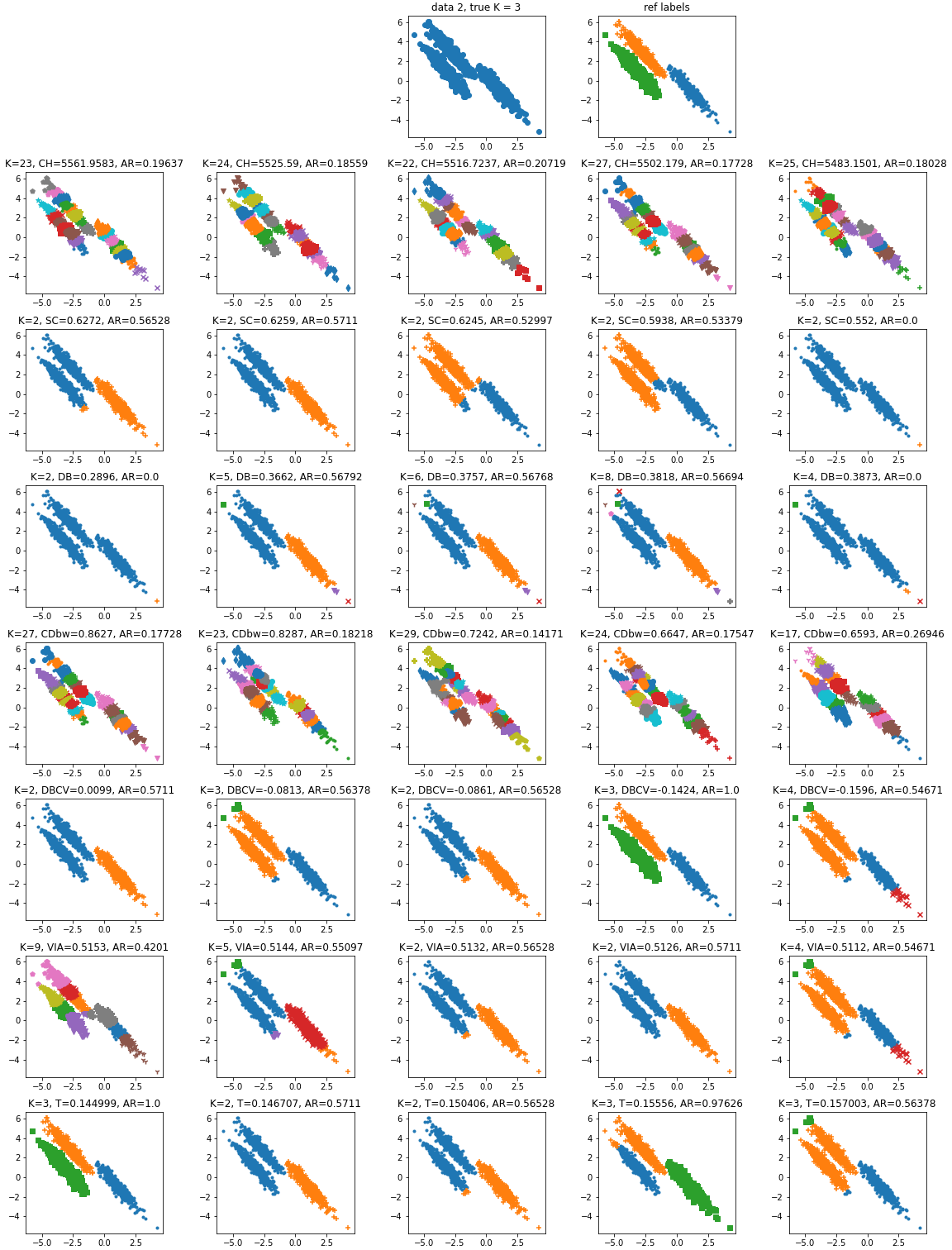}
\end{figure*}
 
\begin{figure*}
	\centering
	\includegraphics[width=1.5\columnwidth]{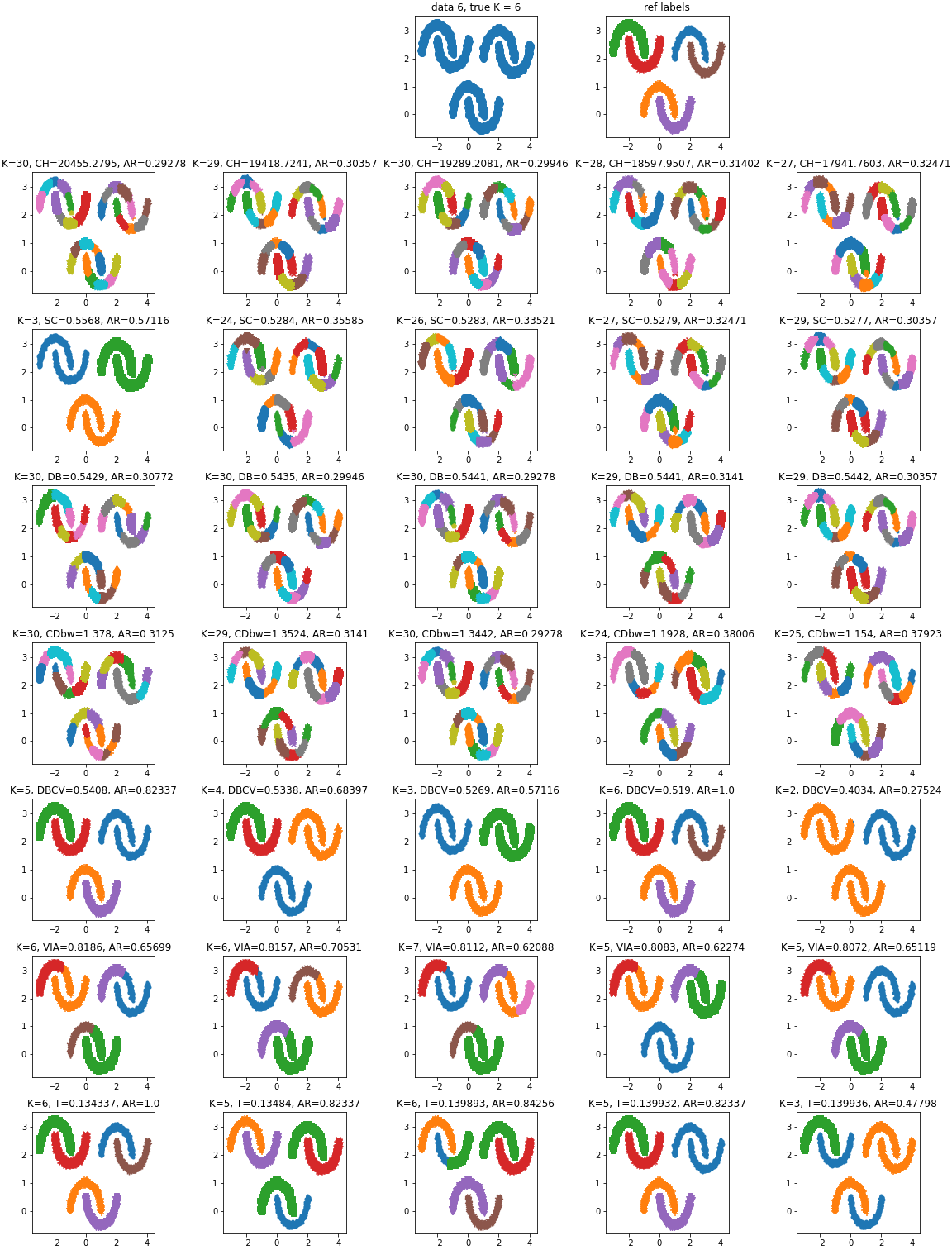}
\end{figure*}

\begin{figure*}
	\centering
	\includegraphics[width=1.5\columnwidth]{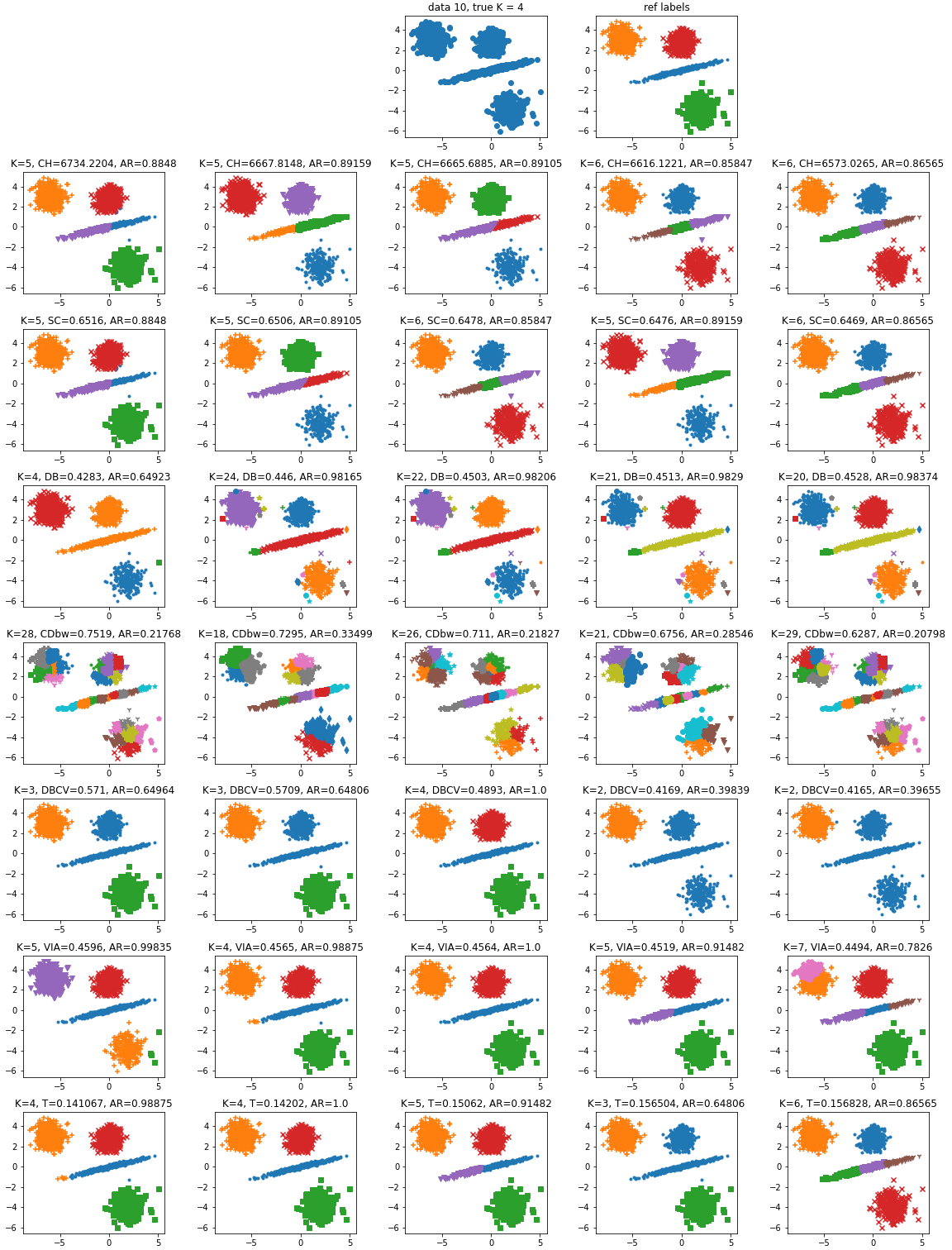}
\end{figure*}

\begin{figure*}
	\centering
	\includegraphics[width=1.5\columnwidth]{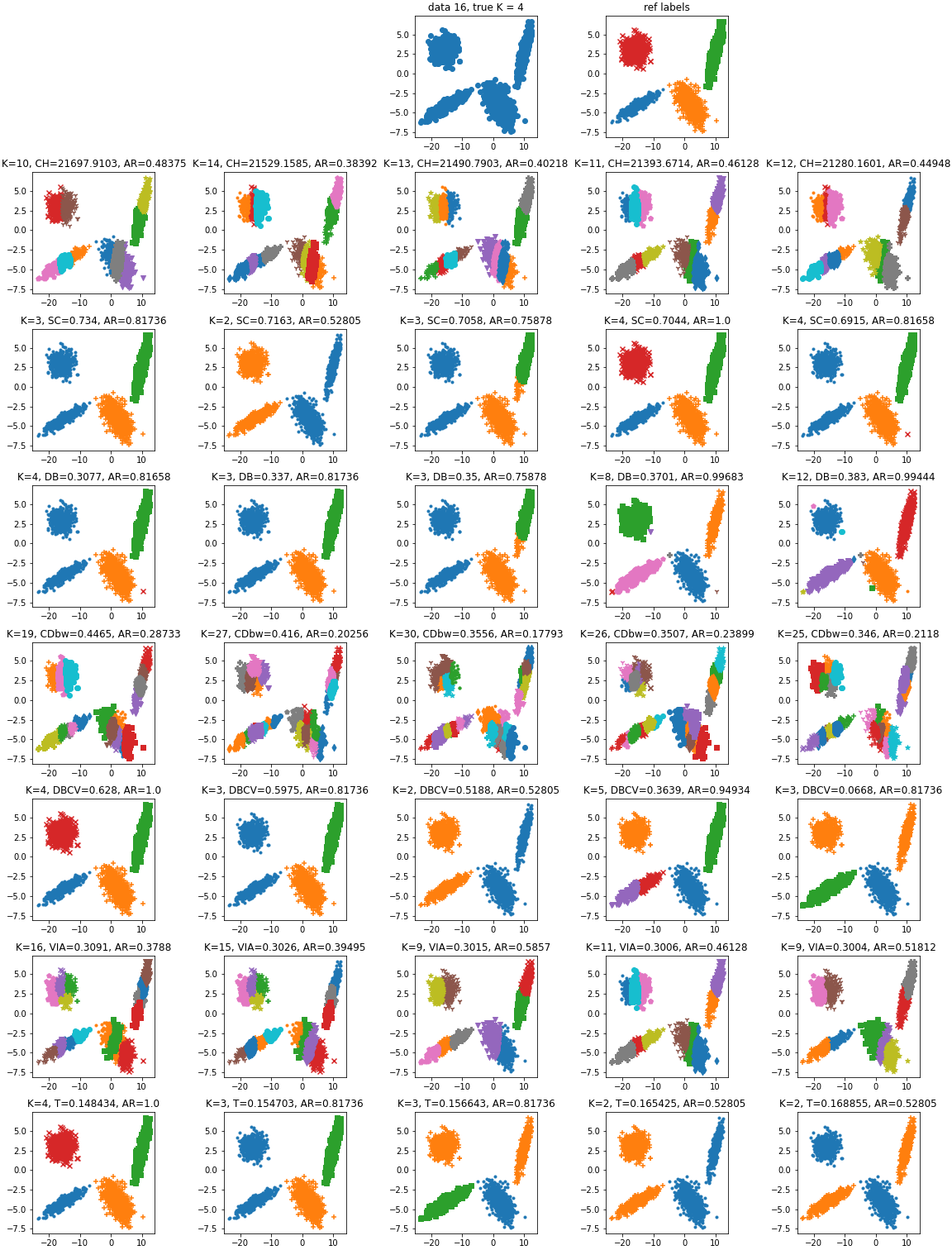}
\end{figure*}

\begin{figure*}
	\centering
	\includegraphics[width=1.5\columnwidth]{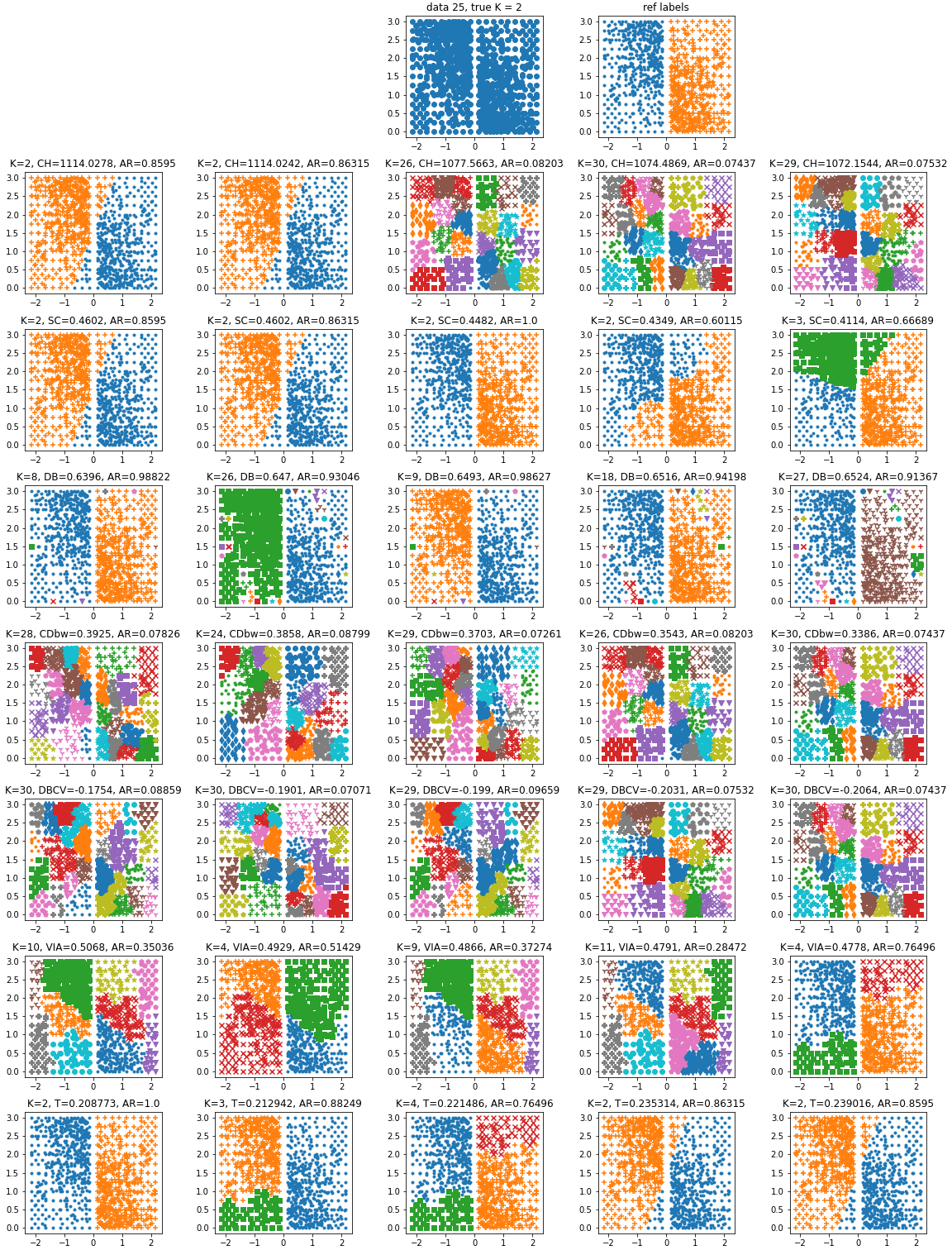}
\end{figure*}

\begin{figure*}
	\centering
	\includegraphics[width=1.5\columnwidth]{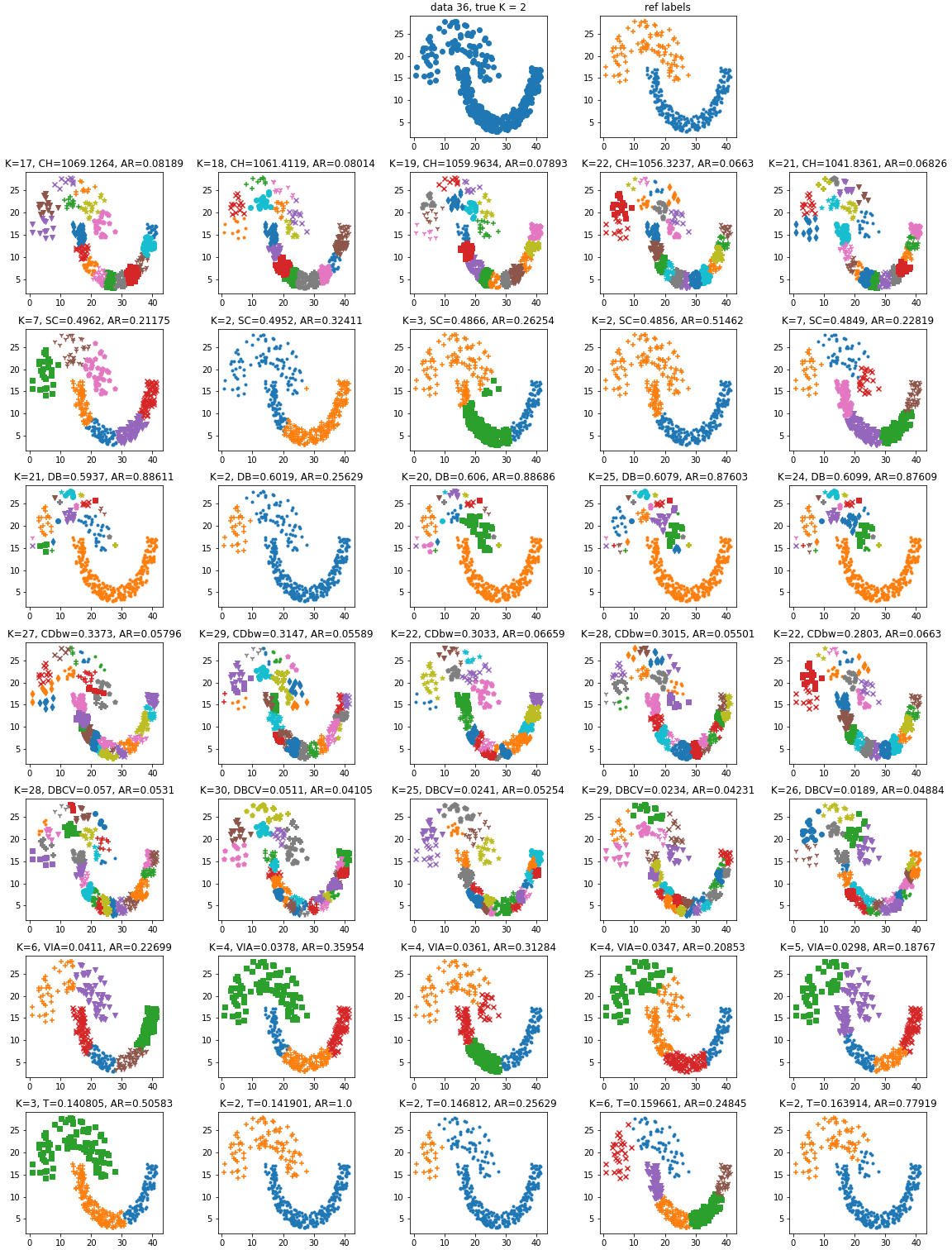}
\end{figure*}

\begin{figure*}
	\centering
	\includegraphics[width=1.5\columnwidth]{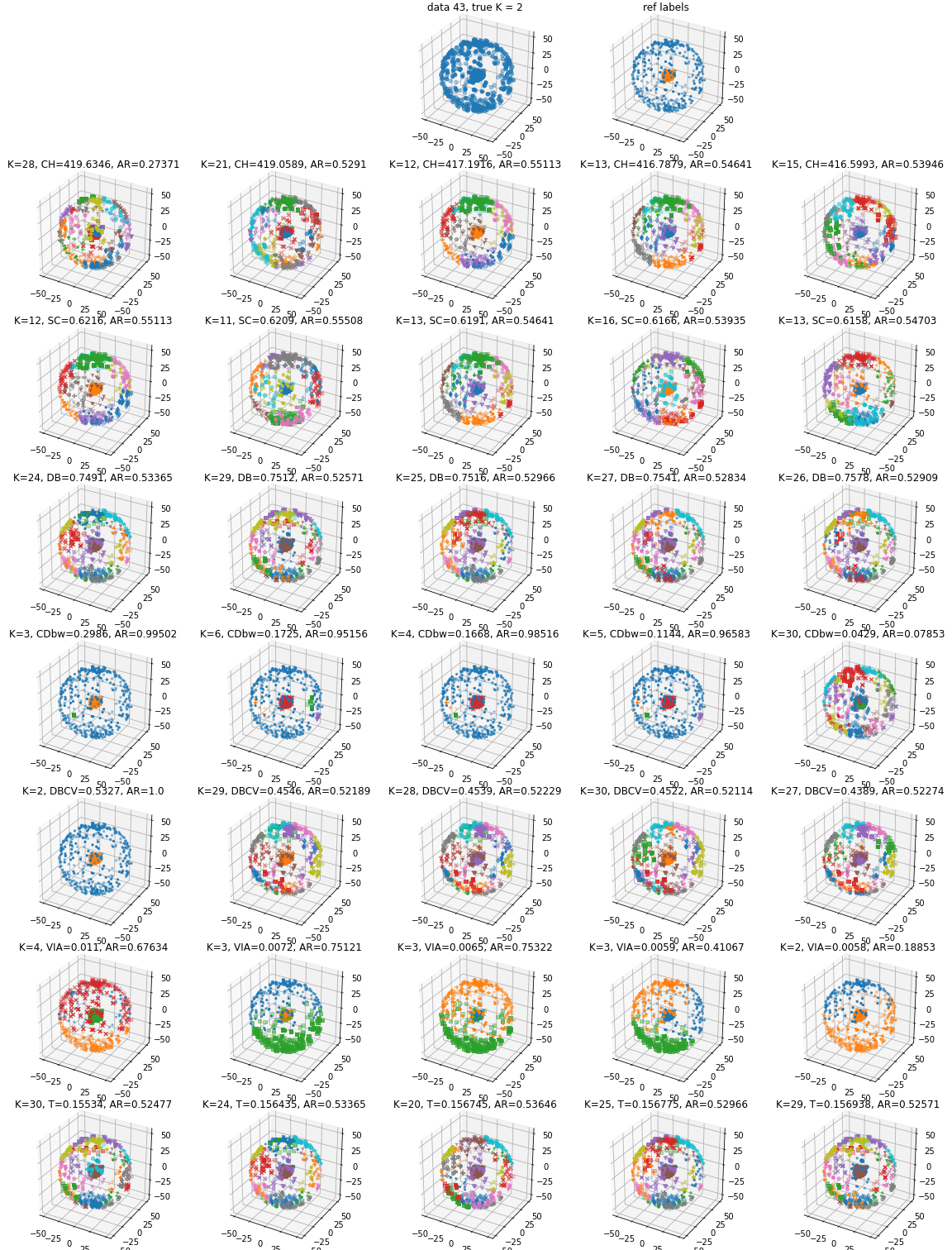}
\end{figure*}

\begin{figure*}
	\centering
	\includegraphics[width=1.5\columnwidth]{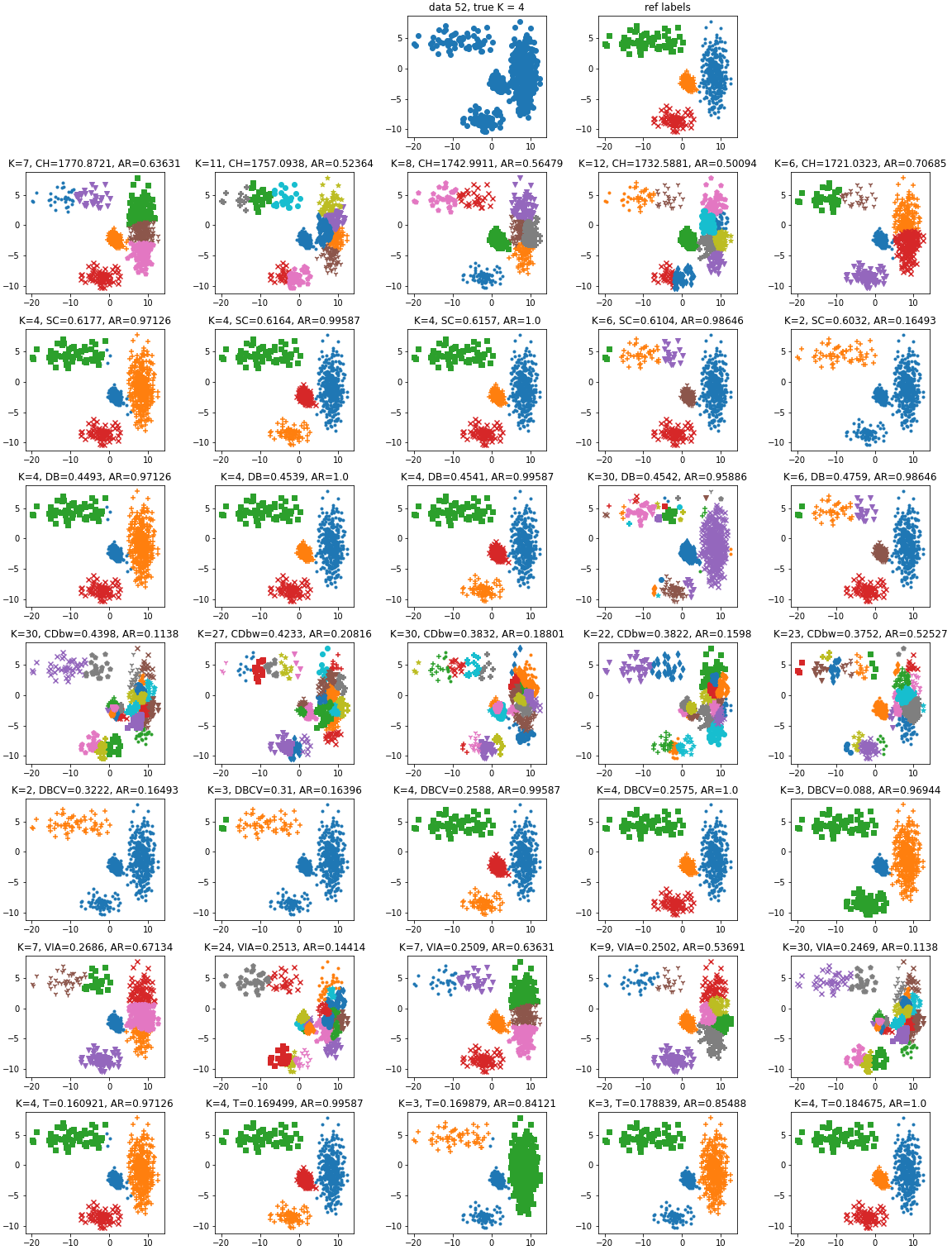}
\end{figure*}

\begin{figure*}
	\centering
	\includegraphics[width=1.5\columnwidth]{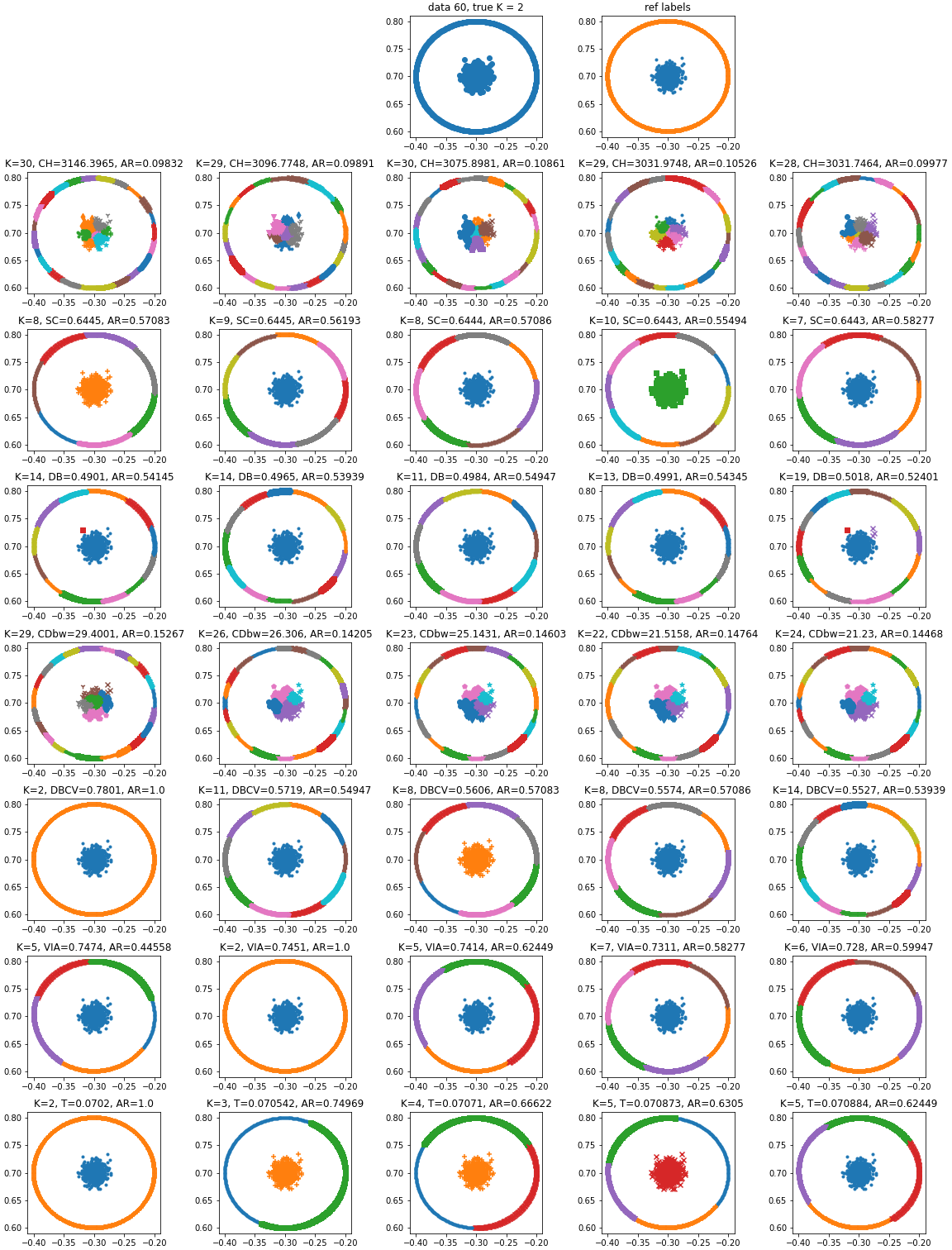}
\end{figure*}

\begin{figure*}
	\centering
	\includegraphics[width=1.5\columnwidth]{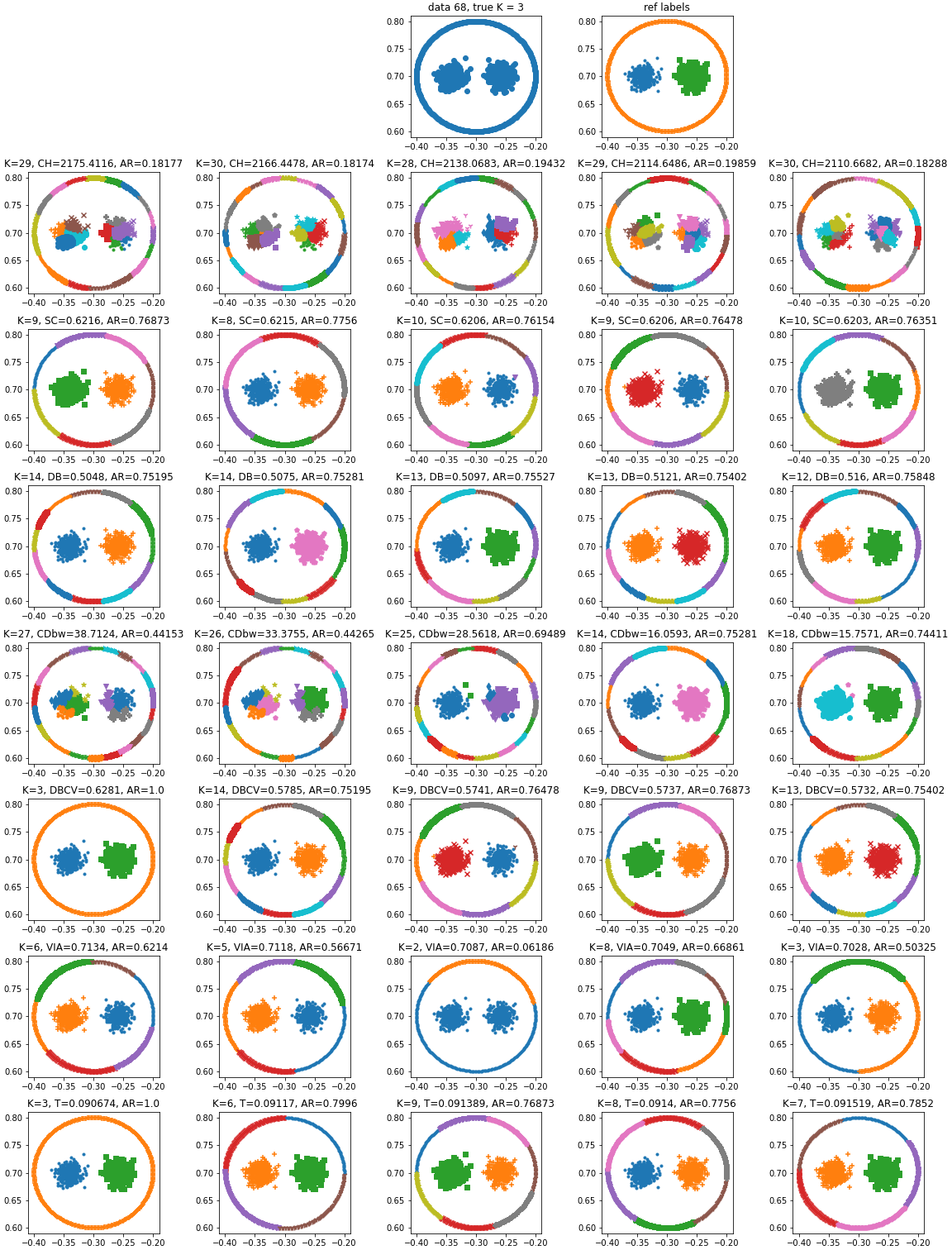}
\end{figure*}

\begin{figure*}
	\centering
	\includegraphics[width=1.5\columnwidth]{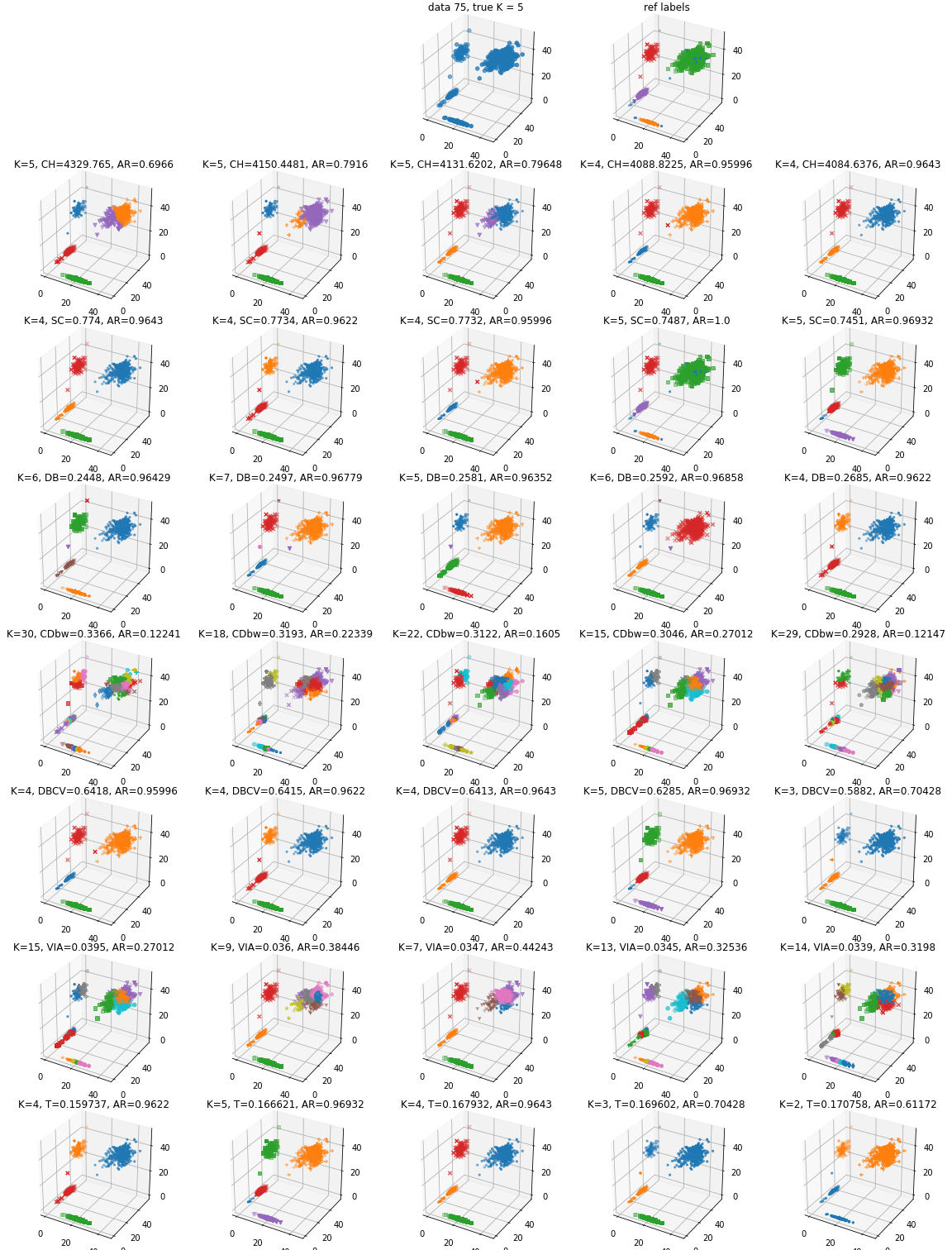}
\end{figure*}

\begin{figure*}
	\centering
	\includegraphics[width=1.5\columnwidth]{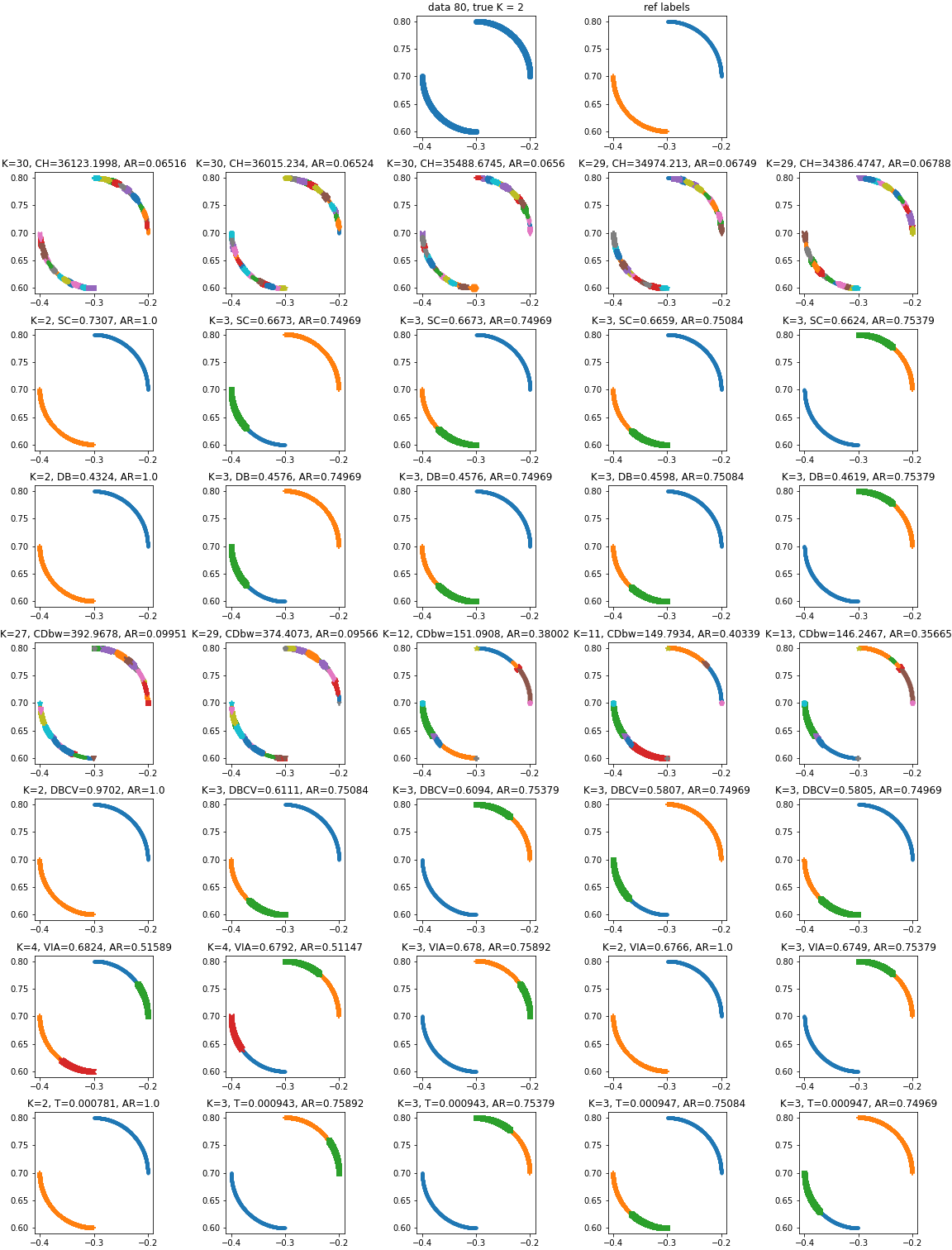}
\end{figure*}

\begin{figure*}
	\centering
	\includegraphics[width=1.5\columnwidth]{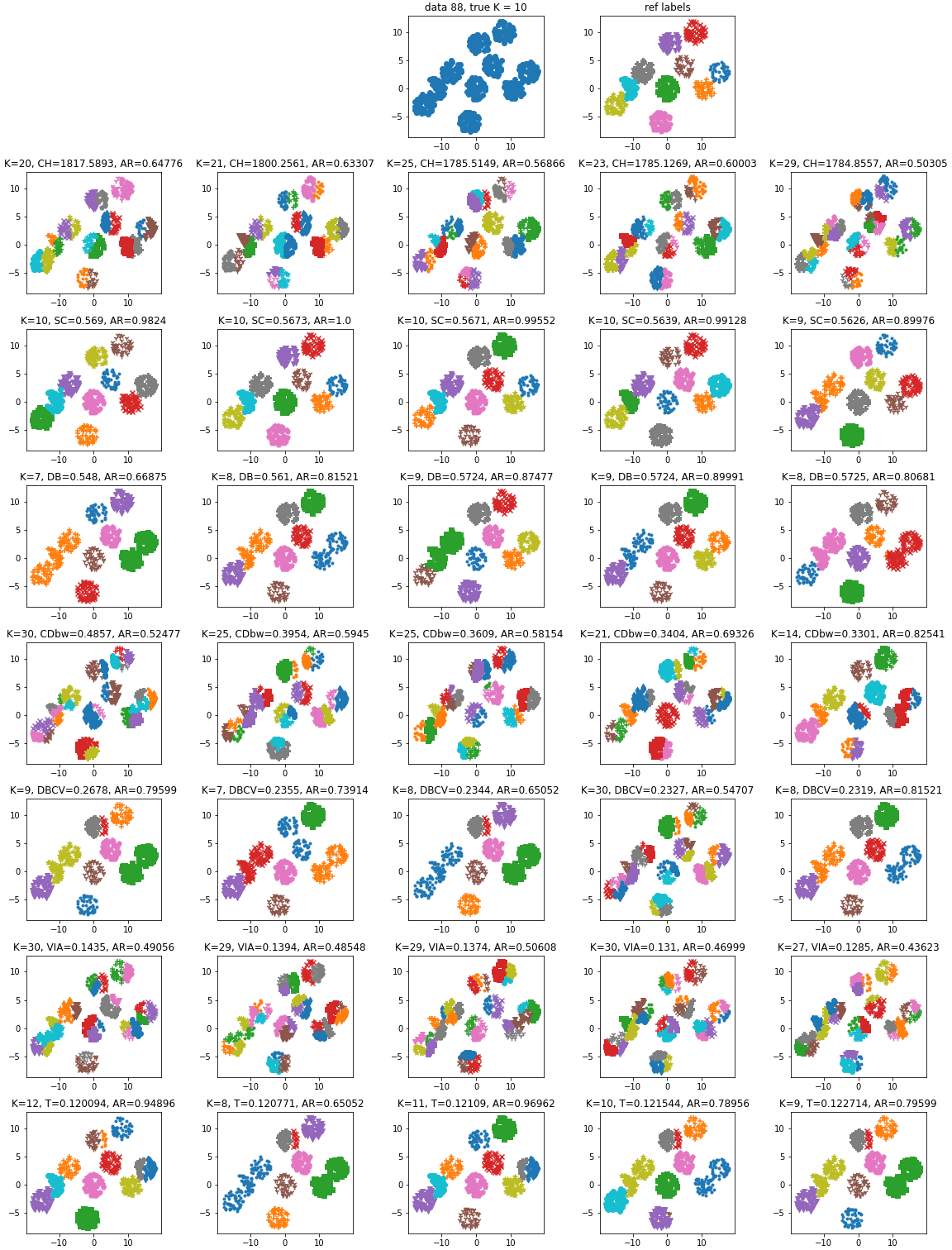}
\end{figure*}

\begin{figure*}
	\centering
	\includegraphics[width=1.5\columnwidth]{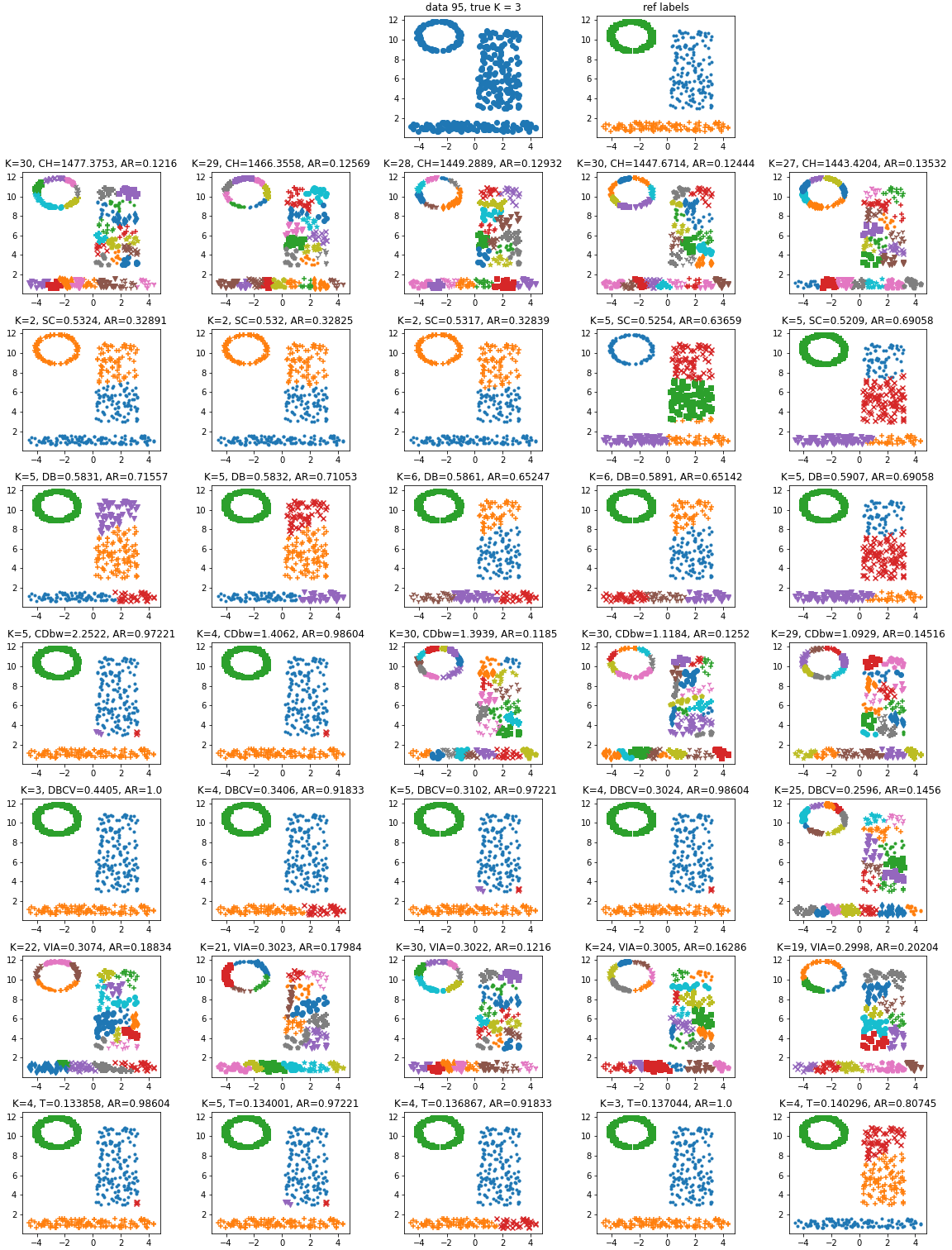}
\end{figure*}

\begin{figure*}
	\centering
	\includegraphics[width=1.5\columnwidth]{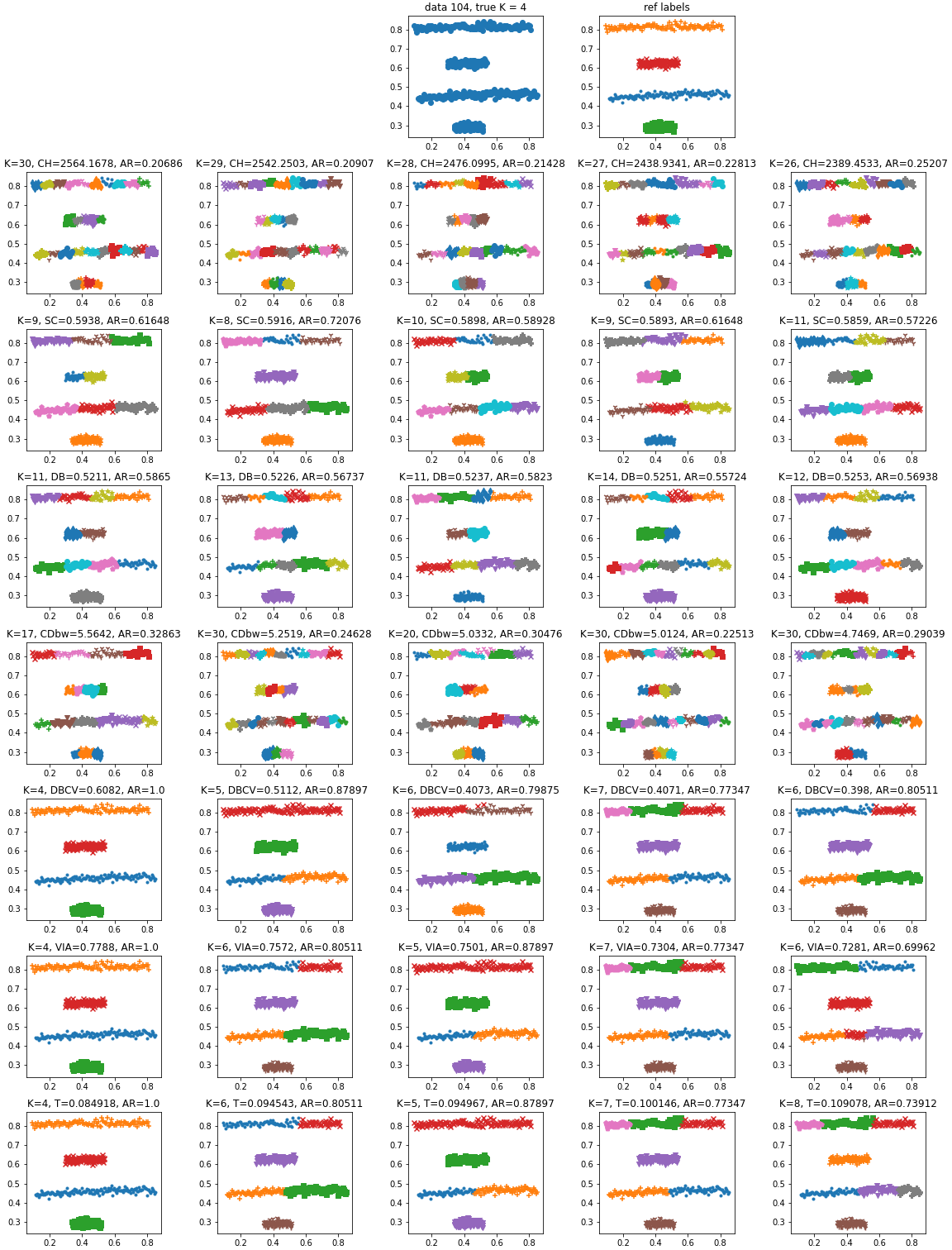}
\end{figure*}

\begin{figure*}
	\centering
	\includegraphics[width=1.5\columnwidth]{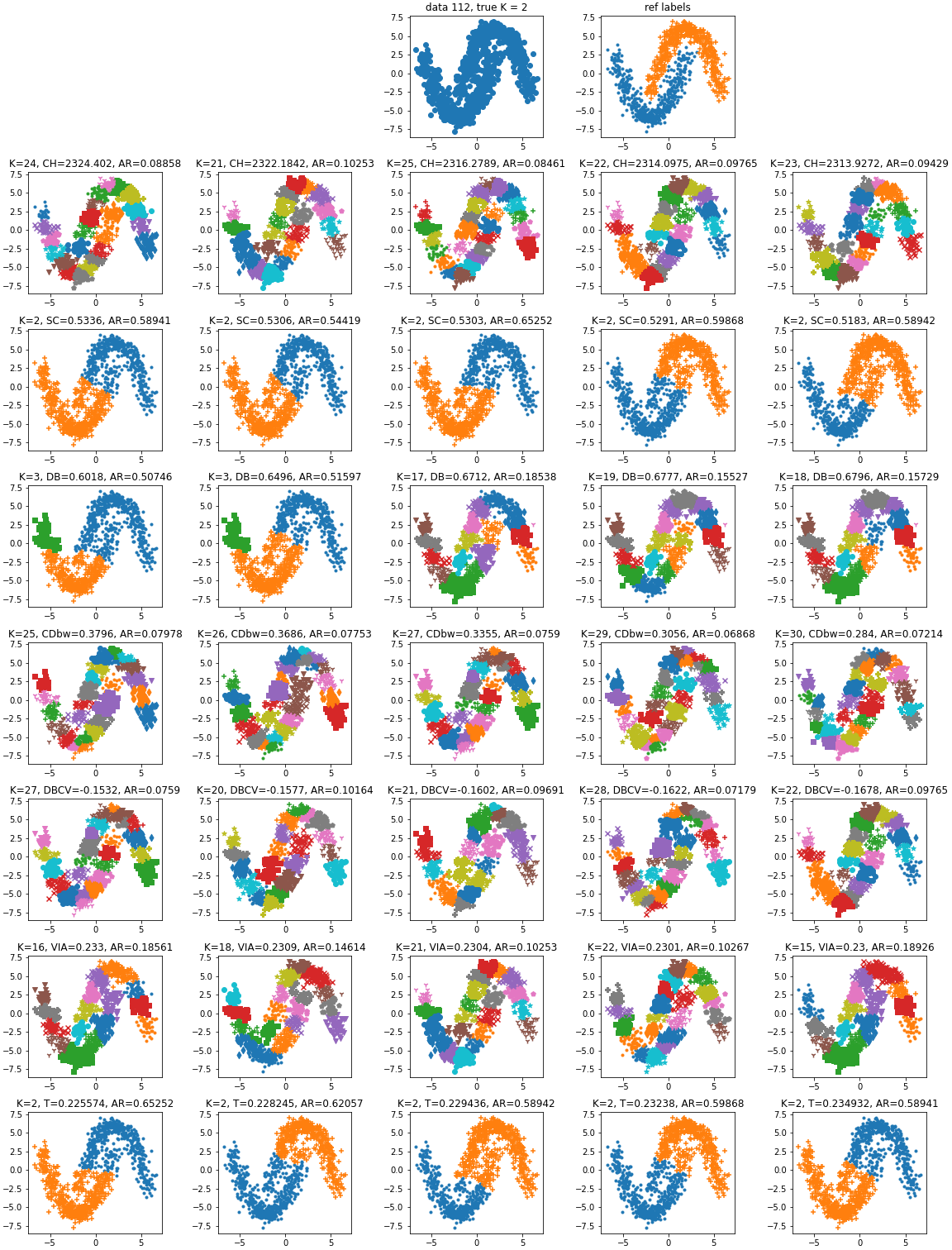}
\end{figure*}

\begin{figure*}
	\centering
	\includegraphics[width=1.5\columnwidth]{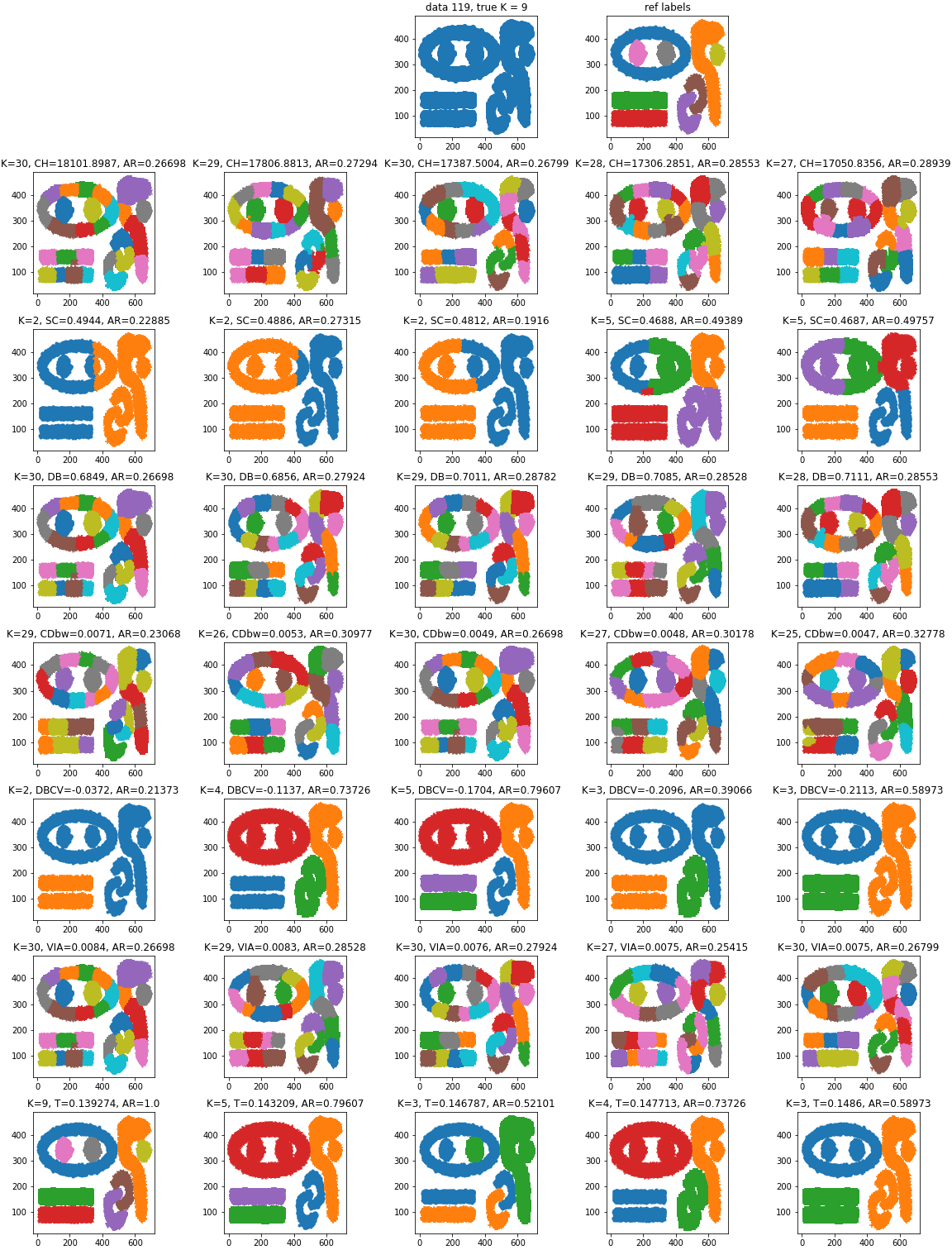}
\end{figure*}

\begin{figure*}
	\centering
	\includegraphics[width=1.5\columnwidth]{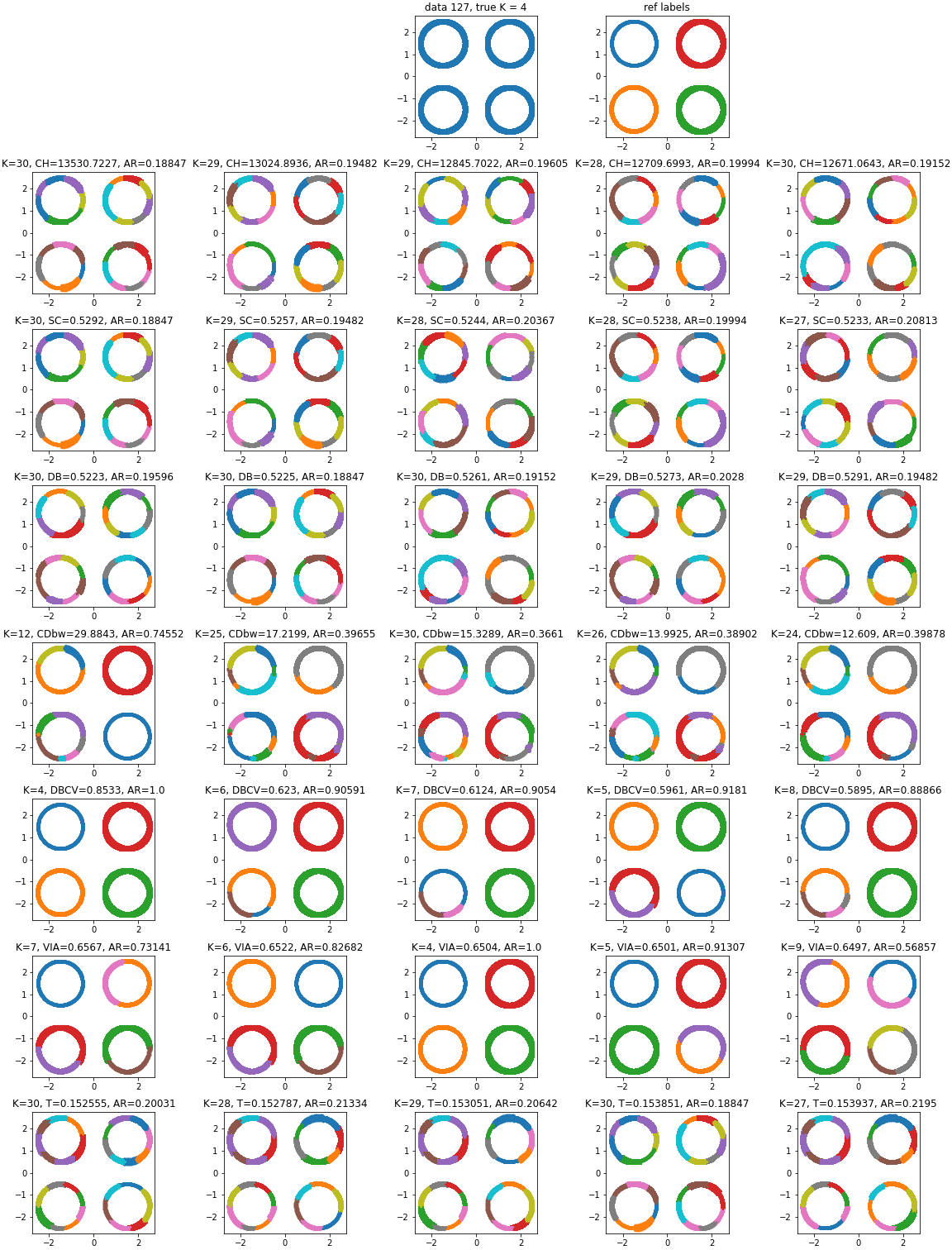}
\end{figure*}

\begin{figure*}
	\centering
	\includegraphics[width=1.5\columnwidth]{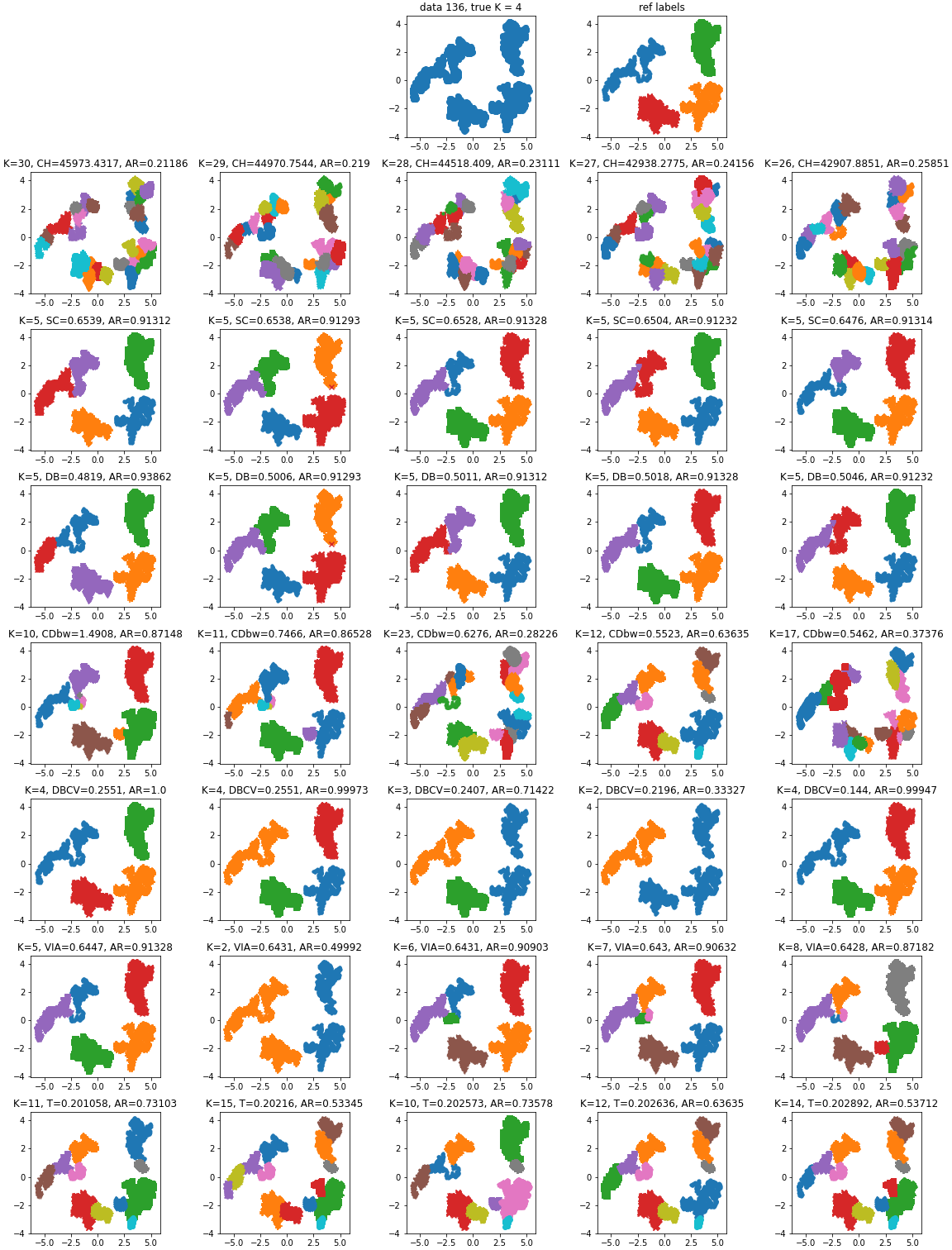}
\end{figure*}

\begin{figure*}
	\centering
	\includegraphics[width=1.5\columnwidth]{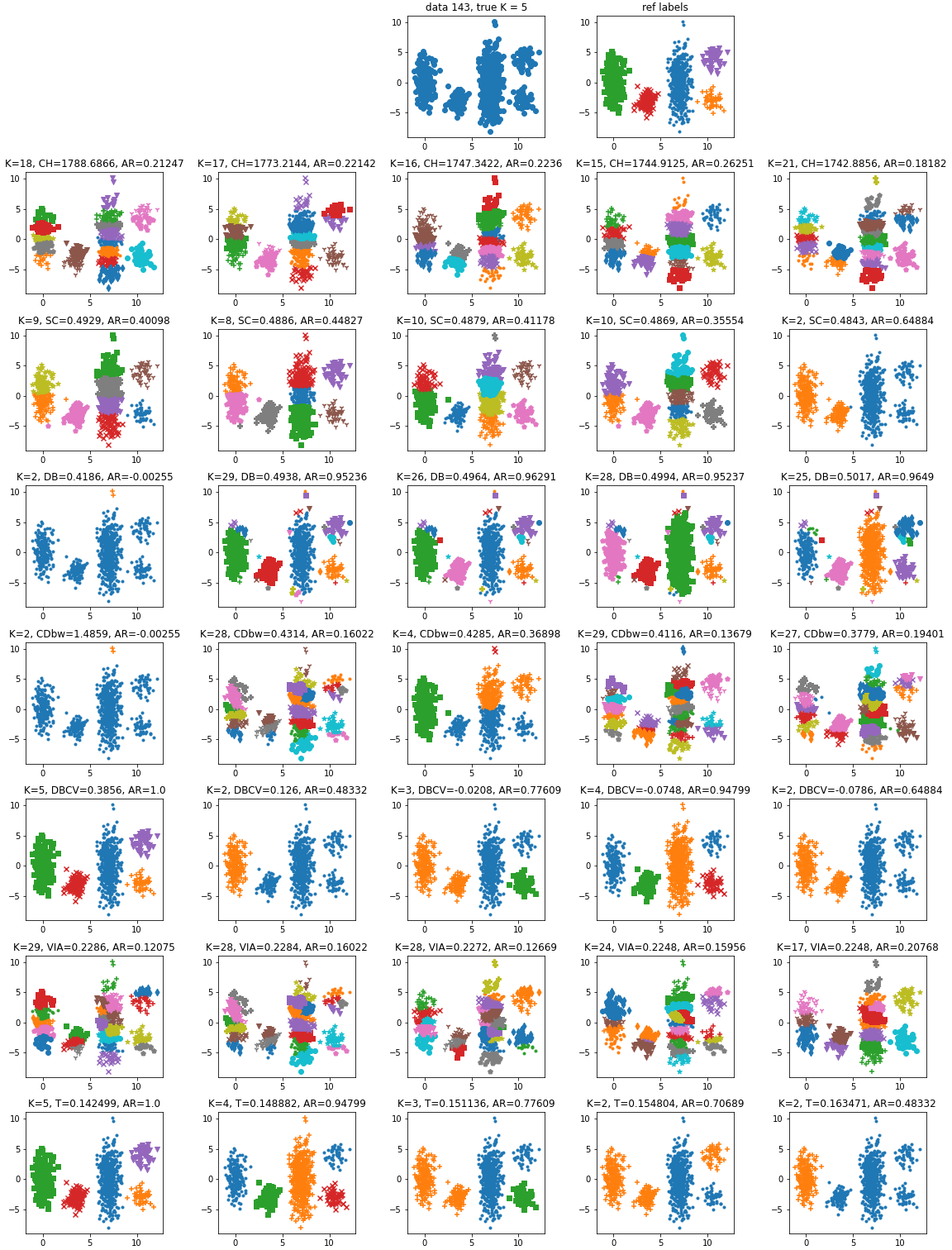}
\end{figure*}

\end{document}